\documentclass[lettersize,journal]{IEEEtran}
\usepackage{amsmath,amsfonts}
\usepackage{algorithmic}
\usepackage{algorithm}
\usepackage{array}
\usepackage[caption=false,font=normalsize,labelfont=sf,textfont=sf]{subfig}
\usepackage{textcomp}
\usepackage{stfloats}
\usepackage{url}
\usepackage{verbatim}
\usepackage{graphicx}
\usepackage{cite}

\usepackage{multirow}
\usepackage{makecell}
\usepackage{soul}
\usepackage[para,online,flushleft]{threeparttable}
\usepackage{booktabs}
\usepackage{ulem}
\usepackage{bm}
\usepackage[inkscapelatex=false]{svg}
\usepackage{nicematrix, tikz}
\usepackage{amssymb}
\usepackage[inkscapelatex=false]{svg}
\usepackage{bm}
\usepackage[para,online,flushleft]{threeparttable}
\usepackage{subfig}
\usepackage{tcolorbox}
\usepackage{graphicx}

\usepackage{algorithm}  
\usepackage{algorithmic}
\usepackage{amsmath}
\floatname{algorithm}{Procedure}

\renewcommand{\emph}[1]{\textit{#1}}
\renewcommand{\underline}[1]{#1}
\begin{document}
	
	\title{MASt3R-Fusion: Integrating Feed-Forward \\  Visual Model with IMU, GNSS for High-Functionality SLAM
	}
	
	\author{Yuxuan Zhou, Xingxing Li\textsuperscript{*}, Shengyu Li, Zhuohao Yan, Chunxi Xia, Shaoquan Feng

		\thanks{This work was supported by the the National Science Fund for
			Distinguished Young Scholars of China (42425401), the National
			Natural Science Foundation of China (423B240), the
			National Key Research and Development Program of China (2023YFB3907100), and the "Sharp Knife" Technology Research Project of Science and Technology Department of Hubei Province (2023BAA025).\textit{(Corresponding author: Xingxing Li.)}}
		\thanks{The authors are with School of Geodesy and Geomatics, Wuhan University, China  (e-mail: xingxingli@whu.edu.cn).}
	}
	
	\markboth{Journal of \LaTeX\ Class Files,~Vol.~14, No.~8, August~2021}%
	{Shell \MakeLowercase{\textit{et al.}}: A Sample Article Using IEEEtran.cls for IEEE Journals}
	
	\IEEEpubid{ }
	
	\maketitle
	
	\begin{abstract}
		Visual SLAM is  a cornerstone technique in robotics, autonomous driving and extended reality (XR), yet classical systems often struggle with low-texture environments, scale ambiguity, and degraded performance under challenging visual conditions. Recent advancements in feed-forward neural network-based pointmap regression have demonstrated the potential to recover high-fidelity 3D scene geometry directly from images, leveraging learned spatial priors to overcome limitations of traditional multi-view geometry methods.  
		However, the widely validated advantages of probabilistic 
		multi-sensor information fusion are often discarded in these 
		pipelines. In this work, we propose \textit{MASt3R-Fusion}, 
		a multi-sensor-assisted visual SLAM framework that tightly integrates feed-forward 
		pointmap regression with complementary sensor information, including inertial
		 measurements and GNSS data. The system introduces Sim(3)-based visual
		  alignment constraints (in the Hessian form) into a universal metric-scale SE(3) factor graph for effective information fusion.
		   A hierarchical factor graph design is developed, which allows both 
		  real-time sliding-window optimization and global optimization with aggressive loop closures, 
		  enabling real-time pose tracking, metric-scale structure perception and globally consistent mapping.
		We evaluate our approach on both public benchmarks and self-collected datasets, 
		demonstrating substantial improvements in accuracy and robustness over existing visual-centered multi-sensor SLAM systems. 
		The code will be released open-source to support reproducibility and further research\footnote{ https://github.com/GREAT-WHU/MASt3R-Fusion}.  
		\end{abstract}
	
	
\begin{figure}[!t]
	\centering
	\includegraphics[width=8.0cm]{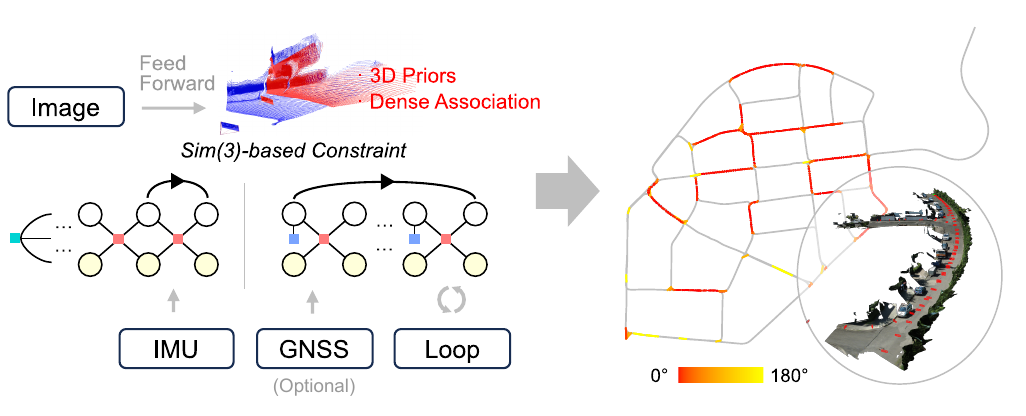}
	\caption{Illustration of the MASt3R-Fusion system. 
	Building upon the  3D perception and data association provided by feed-forward model , 
	this system tightly integrates multi-sensor information (e.g., IMU, GNSS), 
	achieving globally consistent pose estimation and mapping. The heatmap indicates the 
	inter-frame angle of cross-temporal  data association. The system is capable of handling arbitrarily long sequences based on \textbf{8 GB} GPU memory.}
	\label{fig_abstract}
\end{figure}

To overcome these limitations, the integration of deep learning/foundation models has 
emerged as an effective solution\cite{mokssit_deep_2023}. In particular, recent feed-forward neural 
network-based approaches for visual geometry estimation (e.g., DUSt3R\cite{wang_dust3r_2024}, MASt3R\cite{leroy_grounding_2024}, VGGT\cite{wang_vggt_2025}),
have attracted growing attention. These methods encode the images and decode the latents to directly 
recover 2D-to-3D pointmaps and other geometric information. 
By leveraging large-scale data 
to learn the spatial priors, they alleviate many of the degeneracies encountered in traditional 
multi-view stereo (MVS) and SLAM systems, which greatly inspire recent SLAM implementations\cite{murai_mast3r-slam_2025,maggio_vggt-slam_2025}.

Despite significant progress, vision-based methods still suffer from inherent 
limitations, especially regarding scale consistency and performance degradation in visually 
deprived environments. To address this, a promising direction is to  integrate the
 novel visual paradigms 
with complementary sensors, 
such as inertial measurement units (IMUs) and global navigation satellite system (GNSS), 
thus to exploit the respective strengths of traditional multi-sensor 
fusion and learning-based spatial priors.
The integration of multi-sensor information in visual SLAM systems has demonstrated 
 remarkable effectiveness\cite{leutenegger_keyframe-based_2015,qin_vins-fusion_2020,li_continuous_2022}; 
However, how to flexibly apply this approach to emerging vision-based methods 
remains to be explored.

In this work, we propose \textit{MASt3R-Fusion}, 
a framework that integrates multi-sensor information with the 
feed-forward pointmap regression paradigm, facilitating high-functionality  SLAM that supports real-time state estimation, 
complete-view structure perception and
globally consistent mapping.

The contributions of this  work are listed as follows:
\begin{enumerate}
	\item We propose a framework to fuse feed-forward point-map regression paradigm with multi-sensor information in a tight way,
	bridging Sim(3)-based visual alignment constraints 
	with metric-scale SE(3) states and factors.
	\item Based on the feed-forward model, we develop a real-time 
  visual-inertial SLAM system that supports metric-scale pose estimation 
  and dense scene structure perception.
	\item A globally consistent SLAM system that leverages loop closure and GNSS information is developed, in which
	geometry-based loop closure candidate filtering and full-information iterative optimization are utilized.
	\item Both public and self-made datasets are employed to comprehensively evaluate the system performance.
	\item  The code is made open-source to benefit the community.
\end{enumerate}

\section{Related Work}
\subsection{Feed-Forward Pointmap Regression}

Since the introduction of DUSt3R\cite{wang_dust3r_2024}, the feed-forward pointmap regression paradigm
  has emerged as a prominent research focus in the fields of 3D vision 
  and SLAM. This paradigm enables end-to-end neural networks to directly 
  regress 3D point clouds (pointmaps) from image pairs, 
  thereby overcoming the limitations of traditional multi-view 
  geometry methods, which typically rely on explicit camera 
  parameters and iterative optimization.

Building upon DUST3R, MASt3R\cite{leroy_grounding_2024} introduces several key innovations that address critical bottlenecks,
including a descriptor head for refined matching and the ability of metric scale inference. Recent improvements to the feed-forward model paradigm include 
multi-view extensions\cite{wang_3d_2024,wang_vggt_2025}, dynamic scenes\cite{zhang_monst3r_2024}, 
permutation equivariance\cite{wang_pi_2025}, prior information assistance\cite{jang_pow3r_2025}, and so on.

\subsection{Deep Visual SLAM}
For visual SLAM systems, deep learning has introduced 
transformative changes beyond the classical 
implementations\cite{campos_orb-slam3_2021}. These advancements span 
both modular improvements, such as enhanced feature 
matching\cite{detone_superpoint_2018,li_dxslam_2020}, 
image retrieval\cite{zhan_sfm_2025}, depth estimation\cite{tateno_cnn-slam_2017,wang_gat-lstm_2025}, and spatial representations 
including object-level SLAM\cite{adkins_obvi-slam_2024,yang_cubeslam_2019}, renderable neural fields\cite{zhou_mod-slam_2024}, and 3D 
Gaussian Splatting (3DGS)\cite{matsuki_gaussian_2024}. In addition, end-to-end 
approaches 
have also been explored, where images are directly used as input 
to estimate outputs such as camera poses and depth maps. Typically, 
DROID-SLAM\cite{teed_droid-slam_2021} achieves an end-to-end SLAM framework by 
combining dense optical flow with a differentiable dense 
bundle adjustment (DBA) module. Notably, it retains probabilistic 
geometric estimation, resulting in high accuracy and robustness.

More recent efforts have focused on incorporating feed-forward neural networks into SLAM pipelines. 
MASt3R-SLAM\cite{murai_mast3r-slam_2025} 
builds upon MASt3R's two-view geometry to construct an incremental factor graph, 
enabling a scalable SLAM system with high-fidelity scene reconstruction. 
SLAM3R\cite{liu_slam3r_2024} introduces local mapping and global alignment modules, resulting in improved 
multi-frame scene reconstruction performance.
VGGT-SLAM\cite{maggio_vggt-slam_2025}  divides sequences into overlapping submaps and globally aligning them 
on the SL(4) manifold to better handle projective ambiguity and long video sequences.

\begin{figure}[!t]
\centering
\includegraphics[width=8.3cm]{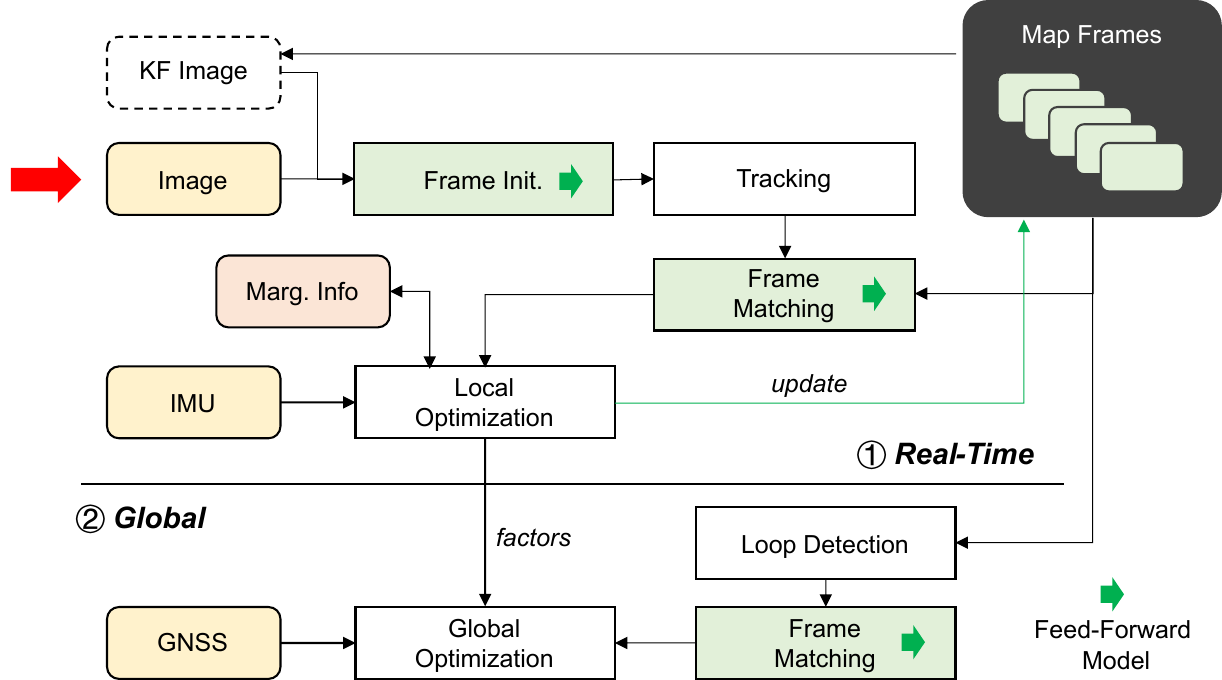}
\caption{Flowchart of the system, consisting of real-time SLAM and global optimization stages.}
\label{fig_flow}
\end{figure}

\subsection{Multi-Sensor-Aided Visual SLAM}

Incorporating multi-sensor information into visual SLAM systems is a 
common approach to enhance system functionality. A representative 
example is the integration of visual and inertial data, 
which enables high-precision, metric-scale pose estimation 
through filter- or optimization-based information fusion\cite{leutenegger_keyframe-based_2015,qin_vins-mono_2018}.
 More advanced methods include photometric-based measurement\cite{von_stumberg_direct_2018}, 
 specialized camera model\cite{jiang_panoramic_2021,wang_pal-slam2_2024}, delayed marginalization\cite{stumberg_dm-vio_2022}, and pose-only modeling\cite{wang_po-kf_2025}.

When it comes to fusing deep learning-based visual methods with 
inertial information,  SL-SLAM\cite{xiao_real-time_2024} and SuperVINS\cite{luo_supervins_2024} incorporate 
learning-based feature modules  to improve the  visual association. 
DBA-Fusion\cite{zhou_dba-fusion_2024} fuses the end-to-end trainable DBA with inertial measurements within a factor graph 
structure.
Additionally, some end-to-end methods\cite{peng_dvi-slam_2024,han_deepvio_2019} directly
feed both images and inertial data into recurrent neural networks for pose estimation.

Apart from IMU, some methods integrate GNSS information for geo-referenced functionality\cite{li_continuous_2022,qin_vins-fusion_2020,cao_gvins_2021}. 
However, it remains a challenge how to 
effectively integrate visual(/-inertial) inforamtion with GNSS for global optimal estimation, as  existing methods either focus solely on real-time navigation 
or rely on pose graph optimization that loses a large amount of visual/inertial information.

In this work, we leverage state-of-the-art feed-forward visual models to formulate  visual alignment constraints and design a system that flexibly integrates multi-sensor information. The resulting framework enables real-time, high-precision pose estimation, dense perception and globally consistent mapping.	
\section{System Overview}
	
The overall system flowchart is illustrated in Fig. \ref{fig_flow}. The framework is mainly composed of two parts: 
real-time SLAM  and global optimization.

In the real-time SLAM stage, input images are processed through a feed-forward neural network to recover 
the pointmaps and achieve further dense matching. Pose tracking is performed by aligning pointmaps between consecutive frames.
A sliding-window factor graph is maintained over a fixed number of keyframes. Sim(3)-based visual constraints 
among co-visible frames are established, while IMU data are 
tightly incorporated. Probabilistic marginalization is applied to maximize information utilization, which enables high-accuracy 
pose estimation and metric-scale structure recovery with low-drift performance.

In the global optimization stage, the information stored from real-time SLAM are processed in a global manner with 
extra loop closures and GNSS information. For loop closure, cross-temporal dense association is achieved through the feed-forward model,
which is boosted by geometry-based candidate selection. 
A global factor graph is employed which incorporates Sim(3)-based visual constraints, IMU pre-integration factors and GNSS position measurements, enabling globally consistent and drift-free pose estimation.

In this system, the feed-forward model introduces two key distinctions from traditional SLAM systems:
\begin{enumerate}
\item \textbf{Powerful 3D priors}, which  enable instant access to the scaleless 3D structure information, leading to significant differences in the construction of visual constraints;
\item \textbf{Aggressive data association}, 
 powered by the 3D awareness, which enables highly effective dense 
 matching to maintain the consistency of visual information
 under extremely large viewpoints.
\end{enumerate}

These features will be integrated into the probabilistic 
multi-senosr fusion framework to maximize their potentials.

\section{Visual Measurement based on Feed-Forward Model}
The typical feed-forward visual model directly regresses  the 3D pointmaps of RGB images in a  common reference frame, and then estimates
 the camera poses through optimization.
This brings about new paradigm for visual-based geometry measurements, 
rather than relying on bundle adjustment (BA) in traditional 
SLAM/structure-from-motion (SfM).

In this work, we generally follow the practice  in MASt3R-SLAM, using feed-forward pointmap regression with
fine-grained matching
and Sim(3)-based  alignment to
construct the visual constraints. Several techniques are used to maximize the geometric accuracy, which would be presented in details.

\subsection{Two-View Feed-Forward Model}
\begin{figure}[!t]
  \centering
  \includegraphics[width=7.4cm]{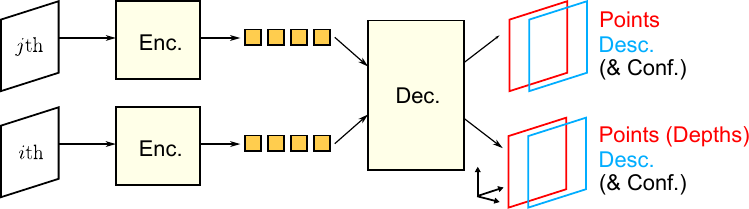}
  \caption{Illustration of the two-view feed-forward model. Two images are encoded into feature tokens, which are then jointly decoded 
  into two 2D-to-3D pointmaps (with $i$ as the common reference frame), together with two descriptor maps.}
  \label{fig_twoview}
\end{figure}

Our work uses MASt3R, a typical two-view feed-forward model to process the images,
which facilitates pointmap regression and dense data association.
The pipeline of  pointmap regression is depicted in Fig. \ref{fig_twoview}. Firstly, images are encoded 
into feature tokens following:
\begin{align}
  \mathbf{F}_i = \mathcal{F}_{\mathrm{enc}}\left(\mathbf{I}_i\right)
\end{align}
where $\mathbf{F}_i$ denotes the  tokens plus the positional encodings. 

For an image pair (denoted as $\mathbf{I}_i$ and $\mathbf{I}_j$),
their tokens are decoded to produce the pointmaps and descriptors:
\begin{align}
  \mathbf{X}^{ij}_i,\mathbf{X}^{ij}_j,\mathbf{D}^{ij}_i,\mathbf{D}^{ij}_j = \mathcal{F}_{\mathrm{dec}}
  \left(\mathbf{F}_i,\mathbf{F}_j\right)
  \label{decode}
\end{align}
where superscript ``$ij$'' indicates the asymmetric  two-view relationship,   $\mathbf{X}^{ij}_i, \mathbf{X}^{ij}_j$ with
$(H \times W,3)$ shapes are 2D-to-3D pointmaps described in the reference frame $i$,  $\mathbf{D}^{ij}_i,\mathbf{D}^{ij}_j$ with
$(H \times W,D)$ shapes
are pixel-level descriptor maps for further  fine-grained association.

Benefiting from large-scale training and the learning capability of transformers, 
the pointmap regression paradigm has been 
demonstrated to be highly effective. Essentially, the model integrates 3D spatial geometric 
priors with two-view stereo to produce reliable structure information 
under diverse conditions.

As the regressed pointmaps are described in the unified reference frame $i$, 
dense matching based on ray proximity can be achieved through 
the following optimization:
\begin{align}
\hat{\mathbf{u}}^i_j =  \arg\min_{\mathbf{u}^i_j}\left\Vert\frac{\mathbf{X}^{ij}_i[\mathbf{u}^i_j]}{\left\Vert\mathbf{X}^{ij}_i[\mathbf{u}^i_j]
\right\Vert} - \frac{\mathbf{X}^{ij}_j}{\left\Vert\mathbf{X}^{ij}_j\right\Vert}\right\Vert^2
\end{align}
where $\mathbf{u}^i_j$ with $(H\times W,2)$ shape denotes the 2D 
coordinates on image $i$ associated with the pixel grids 
on image $j$,  ``$\left[\cdot \right]$'' denotes retrieving the value at 
the corresponding 2D index of the pointmap. Here, 
bilinear interpolation is employed for the retrieval, and the gradient map of $\mathbf{X}^{ij}_i$ is used for efficient optimization.
After the optimization based on ray proximity, point correspondences with large depth residuals
are masked as invalid, which helps eliminate dynamic objects with the awareness of the 3D structure provided by 
the feed-forward model.

As pointed by MASt3R\cite{leroy_grounding_2024}, geometry-based matching can be further improved by taking into 
account  feature-based refinement. To achieve this, the descriptor 
maps are used, following:
\begin{align}
(\hat{\mathbf{u}}^i_j)' =  \arg\max_{\mathbf{u}^i_j} d\left(\mathbf{D}^{ij}_i[\mathbf{u}^i_j],\mathbf{D}^{ij}_j\right)
\end{align}
where $d(\cdot,\cdot)$ denotes the dot-product. As the channel 
number of $\mathbf{D}$ is large and the descriptor map lacks smoothness, 
a neighborhood 
searching (taking $\hat{\mathbf{u}}^i_j$ as initials),  
rather than optimization, is utilized to obtain 
$(\hat{\mathbf{u}}^i_j)'$. 

To achieve sub-pixel data association,
we asymmetrically upsample $\mathbf{D}^{ij}_j$ through 
bilinear interpolation:
\begin{align}
\mathbf{D}^{ij}_{j,\mathrm{up}} = \text{Upsample}_{\text{bilinear}}(\mathbf{D}^{ij}_j, \; 
\text{scale}=4)
\end{align}
then perform the refinement around $(\hat{\mathbf{u}}^i_j)'$, 
following:
\begin{align}
(\hat{\mathbf{u}}^i_j)'' =  \arg\max_{\mathbf{u}^i_j} d\left(\mathbf{D}^{ij}_{j,\mathrm{up}}[\mathbf{u}^i_j],\mathbf{D}^{ij}_i\right)
\end{align}
which contributes to better accuracy for SLAM tasks. 





\subsection{Visual Constraint based on Pointmap Alignment}

Considering multiple image frames, we maintain both the 
pointmaps and their associated poses, represented as:
\begin{align}
\mathcal{K} = \left\{\left(\mathbf{X}_i,\mathbf{S}_i\right) \big| i=0,1,\cdots,K\right\},
\end{align}
where $\mathbf{X}_i$ denotes the pointmap of frame $i$, and
\begin{align}
\mathbf{S}_i = \begin{bmatrix}
s\mathbf{R} & \mathbf{t}\\
0 & 1
\end{bmatrix} \in \mathrm{Sim}(3)
\end{align}
is the camera-to-world Sim(3) transformation, $s\in\mathbb{R}^+$ denotes the scale, $\mathbf{R}\in \mathrm{SO}(3)$ denotes 
the rotation, and $\mathbf{t}\in\mathbb{R}^3$ denotes the translation.
 Essentially, $\mathbf{S}_i$ first scales the points in the camera frame to a unified scale and then 
 transforms them into the world  frame.

The self-referenced pointmap $\mathbf{X}_i$ is initialized using 
the two-view feed-forward results from Eq. (\ref{decode}), following:
\begin{align}
\mathbf{X}_i \leftarrow \mathbf{X}^{ij}_i
\end{align}
in which  frame $j$ is generally the last keyframe.

Specifically, assuming known camera intrinsics, 
the 3D points can be constrained to lie along their corresponding rays:
\begin{align}
\mathbf{X}_i \leftarrow \pi^{-1}\left(\mathbf{u}_i,\left(\mathbf{X}_i\right)_z\right)
\end{align}
where $\mathbf{u}_i$ are the grid-like pixel coordinates, 
$\left(\mathbf{X}_i\right)_z$ is the depth vector, $\pi(\cdot)$ is the camera projection model, 
whose inverse operation
backprojects 2D pixels to 3D points.  
It is noted that, although feed-forward reconstruction is  considered advantageous in that it obviates the requirement for camera parameters,
in practice we found that camera distortions can introduce 
considerable instability. Therefore, we mainly consider the case where the camera parameters are known.

\begin{figure}[!t]
  \centering
  \includegraphics[width=6.4cm]{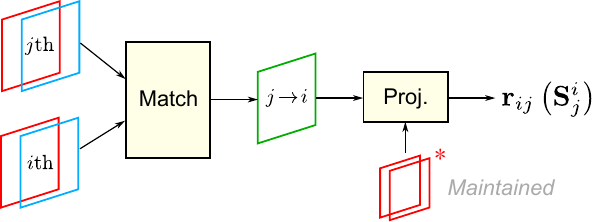}
  \caption{Illustration of matching process and the construction of visual constraints. 
  Note that the matching process is based on the pointmaps from a temporary two-view feed-forward, 
  whereas the projection residuals are constructed using the  pointmaps maintained by the system. }
  \label{fig_matching}
\end{figure}

\begin{figure}[!t]
\centering
\includegraphics[width=8.0cm]{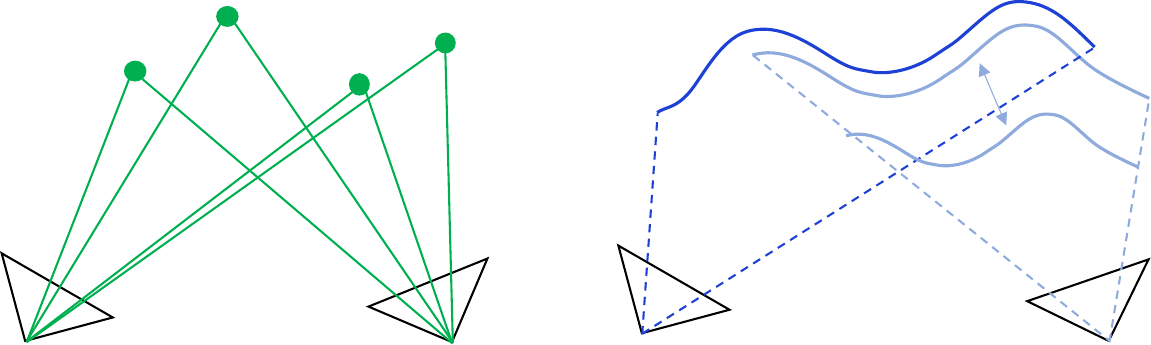}
\caption{Illustration of different forms of visual constraints. \textbf{Left}: bundle adjustment, in which landmarks and 
camera poses are jointly optimized with little prior knowledge. \textbf{Right}: pointmap alignment, 
in which camera poses and pointmap scales are optimized, built upon the knowledge of the scaleless 3D structure.}
\label{fig_alignment}
\end{figure}

Based on the dense association $\mathbf{u}^i_j$ produced by two-view matching, we can align two pointmaps and construct the following
residuals (as shown in Fig. \ref{fig_matching}):
\begin{align}
\mathbf{r}_{ij}\left(\mathbf{S}^i_j \right) = \left[\begin{matrix}
  \mathbf{u}^i_j - \pi\left(\mathbf{S}^i_j\circ \mathbf{X}_j\right) \\
  \left(\mathbf{X}_i\left[\mathbf{u}^i_j\right]\right)_z - 
  \left(\mathbf{S}^i_j\circ \mathbf{X}_j\right)_z
\end{matrix}\right] 
\label{residual}
\end{align}
where $\mathbf{S}^i_j = \mathbf{S}_i^{-1}\circ \mathbf{S}_j$ is the inter-frame relative Sim(3) transformation, ``$\circ$'' denotes the transformation operation. The residuals in the upper part 
resemble the reprojection error with known depth and play a dominant role, 
whereas the residuals in the lower part are primarily used to handle pure rotation 
cases.

The inter-frame transformation can be directly estimated 
based on the above pointmap alignment residuals:
\begin{align}
\hat{\mathbf{S}}^i_j = \arg\min_{\mathbf{S}^i_j}
{   \left\Vert \mathbf{r}_{ij}\left(\mathbf{S}^i_j\right) \right\Vert^2_{\mathbf{Q}_{ij}}
+	\left\Vert \mathbf{r}_{ji}\left(\mathbf{S}^j_i\right) \right\Vert^2_{\mathbf{Q}_{ji}}}
\label{twoframe}
\end{align}
where bi-directional projections are considered, $\mathbf{Q}_{ij}$, $\mathbf{Q}_{ji}$ are the weighting factors which 
consider network-outputted confidence and Huber robust kernel function. This optimization is used for pose tracking of a new frame.

It is noted that, the residual computation and the optimization process are dense due to the 
large number of pixels. Therefore, both the residuals and the Hessian information for each projection are computed on  GPU. To be specific, 
the Hessian inforamtion is computed following:
\begin{align}
\mathbf{r}_{ij} = \mathbf{J}^r_{ij} \boldsymbol{\eta}^i_j\\
\underbrace{(\mathbf{J}^r_{ij})^\top\mathbf{r}_{ij}}_{\mathbf{v}_{ij}} = 
\underbrace{(\mathbf{J}^r_{ij})^\top\mathbf{J}^r_{ij}}_{\mathbf{H}_{ij}} \boldsymbol{\eta}^i_j
\label{hessian}
\end{align}
where $\boldsymbol{\eta}^i_j\in \mathbb{R}^{7}$ is the Lie algebra of $\mathbf{S}^i_j$, $\mathbf{J}^r_{ij}$ is the $\mathbf{r}_{ij}$-to-$\mathbf{S}^i_j$ Jacobian.
After GPU processing, the Hessian-form inforamtion $(\mathbf{H}_{ij},\mathbf{v}_{ij})$ 
are in compact $(7,7)$ and $(7,1)$ shapes  and can be processed efficiently on CPU.

When multiple frames within a neighborhood are considered, the pointmap 
alignment residuals can be constructed for all co-visible image pairs, 
thereby enabling joint estimation of multi-frame poses and scales:
\begin{align}
  \left\{\hat{\mathbf{S}}\right\} = \arg\min_{\left\{{\mathbf{S}}\right\}}
  \sum_{(i,j)\in\mathcal{E}}\left\Vert \mathbf{r}_{ij}\left(\mathbf{S}_i,\mathbf{S}_j\right) \right\Vert^2 _{\mathbf{Q}_{ij}}
\end{align}
where $\mathcal{E}$ is the set of valid two-view projections, corresponding to a directed graph. During the optimization,
relative-to-absolute Jacobians $\mathbf{J}^{ij}_{(i,j)} \in \mathbb{R}^{7\times 14}$ are utilized.


In contrast to BA commonly used in vision-based methods that jointly estimate
point depths and camera poses, 
the above-described visual alignment model does not include point depths, 
and instead builds relatively independent constraints between pairwise images. 
This greatly simplifies the formulation of the  problem, which is made possible 
by the strong capability of the visual foundation model 
to provide accurate estimates of 3D scene structure (up to a certain scale).

For comparison, we revisit the widely used BA technique and the re-prejection error formula,
as shown below:
\begin{align}
\mathbf{r}_{ij}\left(\mathbf{T}^i_j,\mathbf{z}_j \right) = \left[\begin{matrix}
  \mathbf{u}^i_j - \pi\left(\mathbf{T}^i_j\circ \pi^{-1}\left(\mathbf{u}_j,\mathbf{z}_j\right)\right) 
\end{matrix}\right] 
\end{align}
\begin{align}
\left\{\hat{\mathbf{T}},\hat{\mathbf{z}}\right\} = \arg\min_{\left\{\mathbf{T},\mathbf{z}\right\}}
\sum_{i,j\in\mathcal{K}}\left\Vert 
  \mathbf{r}_{ij}\left(\mathbf{T}_i,\mathbf{T}_j,\mathbf{z}_j\right) 
  \right\Vert^2_{\mathbf{Q}_{ij}}
\end{align}
where $\mathbf{z}_j$ are the depths of landmarks in the source frame $j$, $\mathbf{u}_j$ are the pixel coordinates.

It can be seen that when BA constructs visual constraints, 
it relies on the parameterization of a large number of point depths $\mathbf{z}_j$; this greatly increases the scale of the problem and needs careful initialization for good convergence. For better 
efficiency, a two-step optimization can be adopted by eliminating landmarks
first through Schur complement and then constructing pose constraints\cite{triggs_bundle_1999}. 
However, this process involves all frames related to 
$\mathbf{z}_j$, making the construction of visual constraints less flexible. 
These issues no longer exist in pointmap-alignment-based 
visual constraints, which can be easily modeled as pairwise residual terms.

It is noted that, this advantage relies on the assumption that the pointmap
structure is accurate and scale is the only uncertain variable. 
Thanks to the spatial awareness of the feed-forward model, this assumption generally 
holds. Yet, under specific conditions, significant errors may be 
introduced, particularly when the points appear very far in 
image $j$ but very close in image $i$. 
In such cases, the depth uncertainty of the points in image $j$ 
will have a pronounced impact (corresponding to large disparities) in image $i$. 
This issue is not obvious in small-scale scenes, but becomes 
significant in large-scale outdoor environments, typically for the long-time forward-moving scenario.

To mitigate the influence of this uncertainty, we simply apply a mask to downweight the 
residuals in the above mentioned projection process, as shown below:
\begin{align}
\mathbf{r}_{ij}[\text{mask}] = \mathbf{r}_{ij}[\text{mask}] \cdot f_{\text{downweight}}\\
\text{mask} = \left(\mathbf{S}^i_j \circ \mathbf{X}_j\right)_z	< \tau \cdot \left(\mathbf{X}_i\right)_z
\end{align}
where $f_{\text{downweight}}$ is a downweight factor (e.g., 0.1), $\tau$ is  an empirical threshold (e.g., 1.25).

This simple mechanism helps fully leverage 3D prior information  
while mitigating errors caused by unideal conditions, thereby 
contributing to the overall accuracy of the SLAM system.

\section{Real-Time SLAM}

Based on the above paradigm, visual alignment constraints operate  on a frame-to-frame basis, 
involving only a very compact set of associated parameters. 
Beyond the shared common parameters, there exist no functional dependencies among distinct visual constraints. This characteristic enables highly flexible and modular 
construction of factor graphs. Nevertheless, several issues need to be addressed:

\begin{enumerate}
\item Visual constraints are formulated on the Sim(3) group, whereas  measurements from IMU (and  other sensors) are generally defined on SE(3), necessitating appropriate adaptation between these representations.
\item Visual constraints are constructed densely using float32 precision with GPU acceleration. However, float32 precision is insufficient to accurately represent poses at large scales.
\item The dense visual constraint construction depends on the global poses defined in Sim(3). If not carefully handled, the factor graph 
information would change with shifts of the reference frame, leading to more significant linearization error and inflexibility of global/multi-session optimization.
\end{enumerate}

In this section, we construct a  sliding-window-based real-time SLAM pipeline,
during which the above issues are taken into consideration.

\subsection{Isomorphic Group Transformation}

To bridge the gap between Sim(3)-based visual alignment constraints and a more common metric-scale SE(3) factor graph,
we introduce the isomorphic group transformation.

A Sim(3)-based similarity transformation can be described as:
\begin{align}
\mathbf{S}\circ\mathbf{t} =
\underbrace{\begin{bmatrix}
s\mathbf{R} & \mathbf{t}\\
0 & 1
\end{bmatrix}}_{\mathbf{S}\in \mathrm{Sim}(3)}
\begin{bmatrix}
\mathbf{t} \\ 1
\end{bmatrix},
\end{align}

Equivalently, the same transformation can be factorized into an SE(3) transform followed by a scalar scaling:
\begin{align}
\mathbf{T}\circ s \circ \mathbf{t} =
\underbrace{\begin{bmatrix}
\mathbf{R} & \mathbf{t}\\
0 & 1
\end{bmatrix}}_{\mathbf{T}\in SE(3)}
\underbrace{\begin{bmatrix}
s & 0\\
0 & 1
\end{bmatrix}}_{\text{scaling}}
\begin{bmatrix}
\mathbf{t}\\
1
\end{bmatrix}
\end{align}

Thus the similarity group Sim(3) can be equivalently represented in an isomorphic form as the product space $\mathrm{SE}(3)\times \mathbb{R}$:
\begin{align}
\mathbf{S}\in \mathrm{Sim}(3), \quad (\mathbf{T},s)\in \mathrm{SE}(3)\times\mathbb{R}
\end{align}

By introducing Lie algebras and perturbations, we obtain the following relationship:
\begin{align}
\mathbf{S}\boxplus\bm{\eta}
=\left(\mathbf{T}\boxplus\bm{\xi}\right)
\circ\left(s+\delta s\right)
\end{align}
where
\begin{align}
\bm{\eta} = \begin{bmatrix}
\bm{\omega} \\ \bm{\nu} \\ \sigma
\end{bmatrix}_{7\times1},
\quad
\bm{\xi} = \begin{bmatrix}
\bm{\theta} \\ \bm{\tau}
\end{bmatrix}_{6\times1}
\end{align}
with $\bm{\omega},\bm{\theta}\in\mathbb{R}^3$ denoting rotation perturbations, $\bm{\nu},\bm{\tau}\in\mathbb{R}^3$ denoting translation perturbations, and $\sigma,\delta s\in\mathbb{R}$ representing scale perturbations.

On this basis, we can derive the linear transformation relationship between the  Lie algebras:
\begingroup
\setlength{\arraycolsep}{2pt} 
\begin{align}
\begin{bmatrix}
s\mathbf{R} & \mathbf{t} \\ 0 & 1
\end{bmatrix}
\begin{bmatrix}
\bm{\omega}^\wedge + \sigma \mathbf{I} & \bm{\nu} \\ 0 & 1
\end{bmatrix}
=
\begin{bmatrix}
\mathbf{R} & \mathbf{t} \\ 0 & 1
\end{bmatrix}
\begin{bmatrix}
\bm{\theta}^\wedge & \bm{\tau} \\ 0 & 1
\end{bmatrix}
\begin{bmatrix}
s + \delta s & 0 \\ 0 & 1
\end{bmatrix}
\end{align}
\endgroup

This leads to the following compact relation:
\begin{align}
\begin{bmatrix}
\bm{\omega} \\ \bm{\nu} \\ \sigma
\end{bmatrix}
=
\underbrace{
\begin{bmatrix}
1 & & \\
& s\mathbf{I} & \\
& & s
\end{bmatrix}}_{\boldsymbol{\Lambda}}
\begin{bmatrix}
\bm{\theta} \\ \bm{\tau} \\ \delta s
\end{bmatrix}
\end{align}
where $\boldsymbol{\Lambda}$ denotes the diagonal scaling transformation between the two Lie algebra representations.

By introducing this transformation, Sim(3)-based visual constraints can be  consistently applied to SE(3) poses, enabling joint optimization with IMU, GNSS, and other metric-scale sensor information.

\subsection{Multi-Sensor Factor Graph}

In the real-time SLAM stage, we maintain a sliding-window system state as follows:
\begin{align}
\mathcal{X}= \left\{\mathcal{X}_i \mid i\in \mathcal{W} = \left\{K-N+1,\cdots,K\right\}\right\}, \\
\mathcal{X}_i =\left( \mathbf{T}_i, s_i, \mathbf{v}_i, \mathbf{b}_i \right)
\end{align}
where $N$ is the fixed size of the sliding window, and $K$ denotes the index of the latest keyframe, 
$\mathbf{T}_i \in \mathrm{SE}(3)$ is
 the camera pose (rotation and translation) of the $i$-th keyframe,
$s_i \in \mathbb{R}^+$ is the scale parameter,
$\mathbf{v}_i \in \mathbb{R}^3$ is the body velocity in the world frame,
$\mathbf{b}_i = \left(\mathbf{b}_i^g, \mathbf{b}_i^a\right)$ is the IMU bias vector, including gyroscope bias $\mathbf{b}_i^g \in \mathbb{R}^3$ and accelerometer bias $\mathbf{b}_i^a \in \mathbb{R}^3$.
It should be noted that all states are maintained in double precision (float64).

For visual information, a quadric-form factor is utilized, following:
\begin{align}
	\mathbf{E}_v (\mathcal{X}_i,\mathcal{X}_j) =
	\frac{1}{2} l_v\!\left(\mathcal{X}\right)^\top\mathbf{H}_{v,ij}\,l_v\!\left(\mathcal{X}\right) - l_v\!\left(\mathcal{X}\right)^\top \mathbf{v}_{v,ij}\\
	\begin{cases}
	\mathbf{H}_{v,ij} = {\boldsymbol{\Lambda}_{(i,j)}}^\top
{\mathbf{J}^{ij}_{(i,j)}}^\top \mathbf{H}_{ij} \mathbf{J}^{ij}_{(i,j)}\boldsymbol{\Lambda}_{(i,j)}\\
\mathbf{v}_{v,ij} =  {\boldsymbol{\Lambda}_{(i,j)}}^\top
{\mathbf{J}^{ij}_{(i,j)}}^\top \mathbf{v}_{ij}
	\end{cases}
	\label{transformation}
\end{align}
where $(\mathbf{H}_{ij}$, $\mathbf{v}_{ij})$ is the Hessian information of the Sim(3)-based visual constraint (Eq. (\ref{hessian})),
$(\mathbf{H}_{v,ij}$, $\mathbf{v}_{v,ij})$ is the Hessian information after 
relative-to-absolute transformation (through Jacobian $\mathbf{J}^{ij}_{(i,j)}$) and 
isomorphic  group transformation (through $\boldsymbol{\Lambda}_{(i,j)}$),
 $l_v(\cdot)$ is the linear container that transform current state $\mathcal{X}$ 
to the Lie algebras at the  linearization point in Eq. (\ref{residual}) .  

When constructing visual information, we first obtain the relative Sim(3) transformation 
between projection frames using the maintained double-precision (float64) poses and scales. 
The local Hessian information $(\mathbf{H}_{ij},\mathbf{v}_{ij})$
is then computed using Eq. (\ref{residual}) and (\ref{hessian}) in single precision (float32), 
and subsequently transformed via a double-precision linear mapping (Eq. (\ref{transformation})) into 
constraints on the global parameters. By restricting single-precision GPU 
computation within the local scale,
 this strategy mitigates numerical 
instabilities in large-scale scenarios.

Besides, as we use right-hand perturbations, the Jacobians $\mathbf{J}^r_{ij}$ and $\mathbf{J}^{ij}_{(i,j)}$
are only related to the relative transformation $\mathbf{S}^i_j$ and regardless of 
the global reference. This contributes to lower error during marginalization 
and enables more flexible 
use of the factors in global/multi-session optimization.


For IMU information, the classic IMU pre-intergration\cite{forster_imu_2015} is 
used for state prediction and inter-frame dynamic constraints:
\begin{align}
\begin{split}
&\mathbf{r}_{b}\left( \mathcal{X}_k,\mathcal{X}_{k+1}\right)= \\
&\left[\begin{matrix}
{\mathbf{R}^w_{b_k}}^{\top} \left( \mathbf{t}^w_{b_{k+1}}\! -\! \mathbf{t}^w_{b_k}\!+\! \frac{1}{2}\mathbf{g}^w\Delta{t^2_k}\! -\! \mathbf{v}^w_{b_k} \Delta t_k\!\right) -\! \Delta\tilde{\mathbf{t}}^{b_k}_{b_{k+1}}  \\
{\mathbf{R}^w_{b_k}}^{\top} \left( \mathbf{v}^w_{b_{k+1}} + \mathbf{g}^w \Delta t_k - \mathbf{v}^w_{b_k}\right) - \Delta\tilde{\mathbf{v}} ^{b_k}_{b_{k+1}}  \\
\text{Log}\left(\left({\mathbf{R}^w_{b_k}}\right)^{-1}  \mathbf{R}^w_{b_{k+1}}  {\left(\Delta\tilde{\mathbf{R}} ^{b_k}_{b_{k+1}}\right)^{-1}} \right) \\
\mathbf{b}^a_{k+1} - \mathbf{b}^a_{k} \\
\mathbf{b}^g_{k+1} - \mathbf{b}^g_{k}
\end{matrix}\right]
\end{split}
\end{align}
where $\Delta\tilde{\mathbf{t}}^{b_k}_{b_{k+1}}$, $\Delta\tilde{\mathbf{v}}^{b_k}_{b_{k+1}}$, $\Delta\tilde{\mathbf{R}}^{b_k}_{b_{k+1}}$ are the IMU preintegration terms\cite{forster_imu_2015}, $\mathbf{g}^w$ is the gravity vector, $\Delta t_k$ is the time interval.
Considering the difference between the camera pose and the IMU pose, 
the extrinsic parameters need to be taken into account, as shown in 
the following equation:
\begin{align}
	\mathbf{T}^w_{b_i} =\mathbf{T}_i \circ {\mathbf{T}^b_c}^{-1}
\end{align}
where $\mathbf{T}^b_c\in \mathrm{SE}(3)$ is the IMU-camera extrinsic transformation.

The fusion of visual and IMU information forms the fundamental mechanism of visual-inertial odometry (VIO), 
which can provide metric-scale information, low-drift pose estimation, and gravity awareness.
To maximize the performance of this mechanism, probabilistic marginalization needs to be considered. 
To be specific, When the number of states within the sliding window exceeds 
$N$, a local factor graph associated with the oldest pose is constructed. 
The state at the oldest time step is then eliminated via the Schur complement\cite{leutenegger_keyframe-based_2015}, 
thereby yielding the marginalization information $(\mathbf{H}_m,\mathbf{v}_m)$.
 This leads to 
the following quadric-form factor:
\begin{align}
	\mathbf{E}_m (\mathcal{X}) =
	\frac{1}{2} l_m\!\left(\mathcal{X}\right)^\top\mathbf{H}_{m}\,l_m\!\left(\mathcal{X}\right)
	 - l_m\!\left(\mathcal{X}\right)^\top \mathbf{v}_{m}
	\end{align}
	where $l_m(\cdot)$ is the linear container that transform the current states $\mathcal{X}$ 
	to the Lie algebras at the linearization point of marginalization.  
\begin{figure}[!t]
	\centering
	\includegraphics[width=8.0cm]{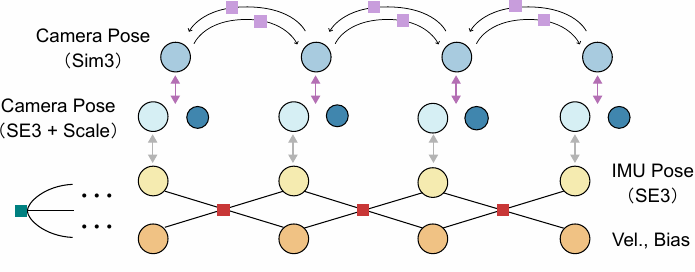}
	\caption{Illustration of the sliding-window factor graph in the real-time SLAM stage. The Sim(3)-to-SE(3) isomorphic group transformation
	and IMU-camera extrinsic transformation are used to bridge the gap between visual constraints and inertial information. Probabilistic marginalization is
	employed for optimal information utilization with controlled problem scale. For 
  clarity, only the adjacent visual constraints are depicted.}
	\label{fig_fgo}
\end{figure}

Thus, the cost function corresponding to the real-time 
visual-inertial SLAM system is described as:
\begin{align}
	\sum_{i \in \mathcal{W}}\left\Vert{\mathbf{r}_{\mathrm{b}}(\mathcal{X}_{i},\mathcal{X}_{i+1})}\right\Vert^2 +
	\sum_{(i,j) \in \mathcal{E}}{\mathbf{E}_{\mathrm{v}}(\mathcal{X}_{i},\mathcal{X}_{j})} + \mathbf{E}_m (\mathcal{X}) 
\end{align}
as is illustrated in Fig. \ref{fig_fgo}.

	With the sliding-window setup, the scale of this optimization problem 
is kept small, and the system is free from the frequent re-computation of older 
projections. In this case, the computational load is mainly 
concentrated in the  feed-forward model. The effective integration of V-I 
information allows the system to achieve high-precision pose/scale estimation using only 
local dense matching, enabling real-time operation on a  laptop with RTX 4080 Mobile GPU. Furthermore, the sliding-window 
design allows the system to handle arbitrarily long data sequences while requiring only \textbf{8 GB} of GPU memory.

\section{Global SLAM}
During the real-time SLAM stage, the V-I information (feature tokens $\left\{\mathbf{F}\right\}$, pointmaps $\left\{\mathbf{X}\right\}$, system states $\left\{\mathcal{X}\right\}$, 
Hessian-form visual information $\left\{\mathbf{H},\mathbf{v}\right\}$ and IMU factors) is 
logged to the file.

In the global SLAM process, based on the pre-processed V-I information and
state estimation from the real-time stage, 
we incorporate loop closure and GNSS data to achieve consistent pose 
estimation and mapping through efficient global optimization.
	\subsection{Loop Closure}
	Loop closure detection is crucial for achieving consistent mapping and drift-free pose 
	estimation in GNSS-denied environments. In particular, the feed-forward visual model provides 
	the fundamental capability for data association across large viewpoints, 
	making it especially worthy of attention.

	First, as proposed in\cite{murai_mast3r-slam_2025}, we can use image feature tokens obtained from the feed-forward 
	encoder to construct an image retrieval system:
	\begin{align}
		\mathbf{d}_{i} = \mathcal{F}_{\mathrm{ret}}
		\left(\mathbf{F}_i\right)
	\end{align}
	\begin{align}
		p^{*}_q = \arg\min_{p}\left\Vert\mathbf{d}_{p} - \mathbf{d}_{q}\right\Vert
	\end{align}
	where $\mathbf{d}_{i}$ is the descriptor vector, $p^{*}_q$ is the retrieved index by query $q$. Considering the 
	need for cross-temporal loop closure, we select the top-10 indices and 
	exclude temporally consecutive frames from them, thereby 
	constructing the loop closure candidate set.

	Benefiting from the strong association 
	capability of the feed-forward model, this system is highly effective to recognize subtle cues
	of a common scene,
	 but also produces 
	a considerable number of false positives. 
	A further dense matching step (see Sect. 4.1) can be used for verification of the loop closure,
	while this can introduce significant extra computational cost.

	To lower the risk caused by false loop closures 
	and to reduce the computational cost of fine-grained verification, 
	an  adaptive filtering mechanism would be  beneficial.
	Building on the low-drift pose estimates
	 provided by VIO, we developed a highly efficient method to eliminate redundant false 
	 detections based on pose uncertainty, which reduces redundancy while preserving as many aggressive loop closure 
	 candidates as possible.

	 Based on the odometry essence of VIO, we simplify the position estimation as a Markov process, 
	 leading to the following expression:
\begin{equation}
	\Delta\hat{\mathbf{t}}^w_{p,q} = \sum_{p}^{q-1}{\Delta\hat{\mathbf{t}}^w_{i,i+1}}
\end{equation}
where $\Delta\hat{\mathbf{t}}^w_{p,q}$ is the relative translation estimation between timestamp $p$ and $q$. 
Considering the practical need of loop closure, we simplify the translation into 2D horizontal movement.

Considering the error of the odometry, we get:
\begin{equation}
    \Delta \hat{\mathbf{t}}^w_{i,i+1} 
    = \Delta \mathbf{t}^w_{i,i+1} + \boldsymbol{\epsilon}_{i,i+1},
\end{equation}
where $\boldsymbol{\epsilon}_{i,i+1}$ is the estimation error, 
 modeled  as a random variable conforming to a certain uncertainty.

By dividing the odometry uncertainty into ``along'' direction (corresponding to scale error) 
and ``cross'' direction (corresponding to heading error), we 
can estimate the translation uncertainty following:
\begin{equation}
    \boldsymbol{\epsilon}_{i,i+1} \sim 
    \mathcal{N}\!\left(\mathbf{0}, \, d^2 \mathbf{Q}\right),
\end{equation}
where $d = \left\|\Delta \hat{\mathbf{t}}^w_{i,i+1}\right\|$ is the translation distance, 
$\mathbf{Q}$ is a direction-and-uncertainty-related ellipsoid, 
following:
\begin{align}
	&\mathbf{Q} = \sigma_d^2 \mathbf{P}_{\parallel} + 
	\sigma_n^2 \mathbf{P}_{\perp}\\
	&\begin{cases}
		\mathbf{P}_{\parallel} &= \mathbf{n}\mathbf{n}^\top\\
		\mathbf{P}_{\perp} &= \mathbf{I} - \mathbf{n}\mathbf{n}^\top\\
		\mathbf{n} &= 
		\Delta \hat{\mathbf{t}}^w_{i,i+1}/\left\|\Delta \hat{\mathbf{t}}^w_{i,i+1}\right\|
	\end{cases}
\end{align}
where $\sigma_d$ is the uncertainty factor related to the scale error of translation,
$\sigma_n$ is the uncertainty factor  related to the direction/heading. These two factors are 
preset parameters considering the normal VIO performance.

Based on the above equations, we can efficiently compute the covariances of $\boldsymbol{\epsilon}_{i,i+1}$, and estimate 
the covariance of  inter-frame translation $\Delta\hat{\mathbf{t}}^w_{p,q}$ by summing up. This operation is fast by utilizing 
vectorization. Afterwards, we turn the uncertainty of translation 
into the uncertainty of inter-frame distance:
\begin{equation}
    \sigma_{p,q} = \sqrt{\frac{{\Delta\hat{\mathbf{t}}^w_{p,q}}^\top \mathbf{Q}_{p,q}\, {\Delta\hat{\mathbf{t}}^w_{p,q}}}
                          {\left\Vert{\Delta\hat{\mathbf{t}}^w_{p,q}}\right\Vert^2}},
\label{sigmapq}
						\end{equation}
where $\mathbf{Q}_{p,q}$ is the estimated covariance of translation $\Delta\hat{\mathbf{t}}^w_{p,q}$.

Equation (\ref{sigmapq})  forms a 2D uncertainty map of inter-frame distances. 
It should be noted that, when predicting the co-visibility, we need to consider not 
only the spatial proximity of camera positions. Due to the presence of large-angle loop closures, 
co-visible frames may observe the same scene while exhibiting significant positional 
differences caused by varying viewpoints. Therefore, it is more important 
to focus on the proximity of observed points. To this end, we approximate the main observation point of a camera frame $i$ by introducing a point 
of interest, defined as:
\begin{equation}
    \mathbf{t}_{\bar{i}} = \left(\mathbf{T}_i \circ
    \begin{bmatrix} 
        0 \\ 0 \\ L 
    \end{bmatrix}\right)_{x,y}
\end{equation}
where $L$ denotes the median scene depth of the corresponding camera frame.

\begin{figure}[!t]
	\centering 
	\includegraphics[width=8.2cm]{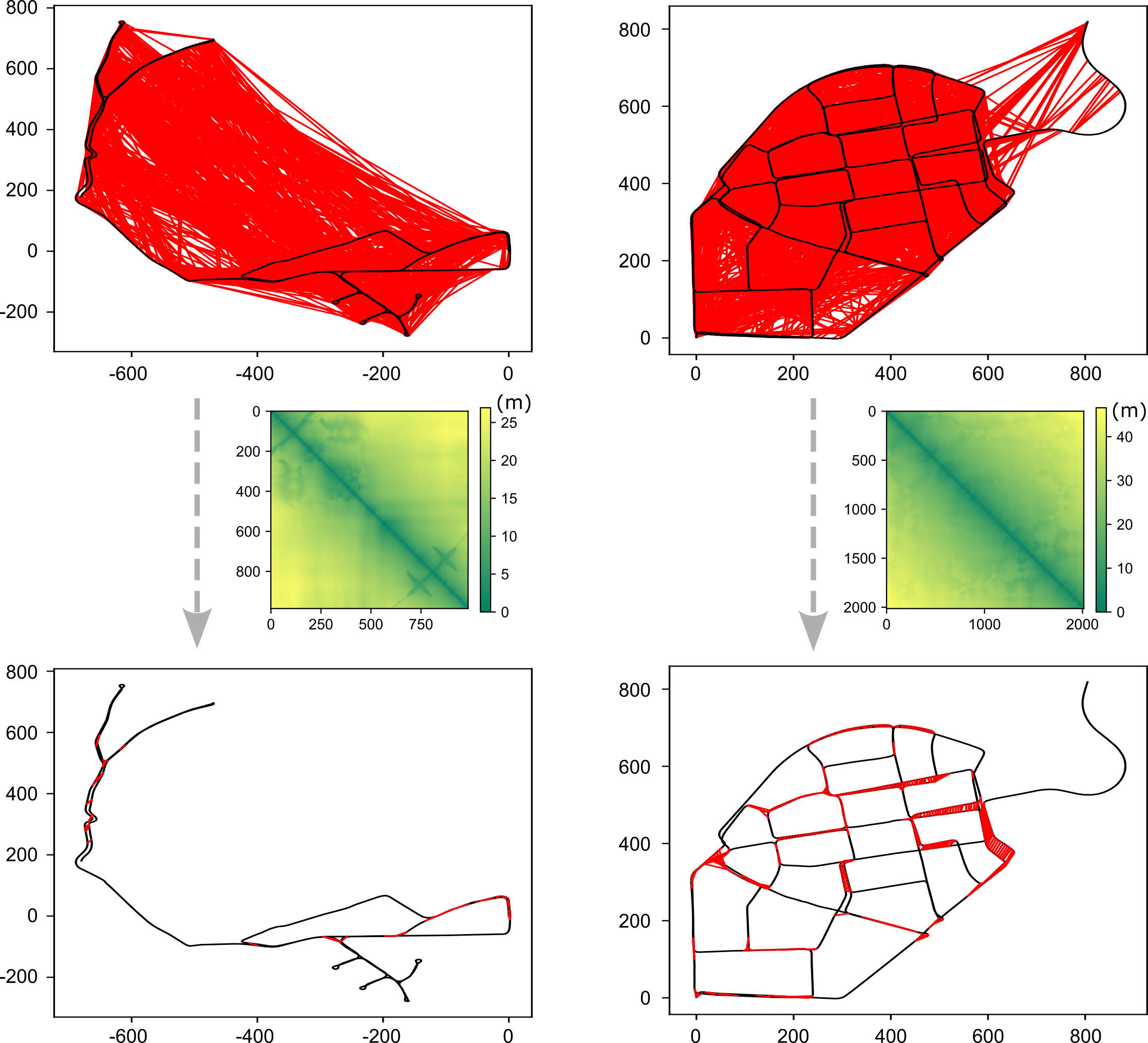}
	\caption{Loop closure candidate filtering based on efficient positional uncertainty evaluation.
	The upper figure illustrates the set of loop closure candidates selected based on the similarity scores and temporal non-contiguity. After filtering with the distance uncertainty matrix 
  $\sigma_{p,q}$ (Eq. \ref{sigmapq}), 
	the lower figure presents the geometrically feasible loop closure candidates.} 
	\label{fig_candidate} 
  \end{figure}

A heuristic criterion for selecting co-visible frames is given as:
\begin{equation}
	d\left(\mathbf{t}_{\bar{q}},\mathbf{t}_{\bar{p}}\right) < L
	\label{crit}
\end{equation}
where $d(\cdot,\cdot)$ is the Euclidean distance.

	The next task is to determine, based on positional uncertainty, whether there is a possibility that two frames satisfy condition (\ref{crit}).
	By approximating the uncertainty of the distance between points 
	$\mathbf{t}_{\bar{p}}$ and $\mathbf{t}_{\bar{q}}$ to that of the distance between points $\mathbf{t}_p$ and $\mathbf{t}_q$, 
	the probability distribution function corresponding to 
  the above condition can be obtained:
\begin{equation}
	P(d\left(\hat{\mathbf{t}}_{\bar{q}},\hat{\mathbf{t}}_{\bar{p}}\right) < L) = 
  \Phi\left(\frac{L - d\left(\hat{\mathbf{t}}_{\bar{q}},\hat{\mathbf{t}}_{\bar{p}}
  \right)}{\sigma_{p,q}}\right)
\end{equation}
where $\Phi(\cdot)$ denotes the standard normal cumulative distribution function (CDF).

On this basis, we take the following criterion to select possible loop closure candidates:
\begin{equation}
	d\left(\hat{\mathbf{t}}_{\bar{q}},\hat{\mathbf{t}}_{\bar{p}}\right) < L + \sigma_{p,q}
\end{equation}
which corresponds to a relatively strict 15.87\% tailed probability to exclude cases that don't satisfy condition (\ref{crit}).

Examples of the above-described candidate filtering are presented in Fig. \ref{fig_candidate}.

\subsection{Global Factor Graph Optimization}

Building upon the above work, the global graph integrates GNSS, loop closure and local constraints for optimal estimation, and leverages sparsity to enable efficient optimization.

For visual loop closure measurements, the relative pose estimate between loop closure 
frame $(i,j)$ can be directly obtained through the projection between two frames (Eq. (\ref{twoframe})), 
which is $\tilde{\mathbf{S}}^i_{j,\mathrm{loop}}$.
With the knowledge of $s_i$ obtained in the real-time SLAM stage, the SE(3) relative pose  can be retrieved, which leads to a metric-scale
relative pose contraint:
\begin{equation}
\mathbf{r}_r (\mathcal{X}_i,\mathcal{X}_j) = {\mathbf{T}_i} \circ {\mathbf{T}_j}^{-1} 
\circ \tilde{\mathbf{T}}^i_{j,\mathrm{loop}}
\label{relative}
\end{equation}
where $\tilde{\mathbf{T}}^i_{j,\mathrm{loop}}$ is the estimated relative pose between the loop closure frames.

For more accurate constraints, we also store the Hessian-form information $(\mathbf{H}_{ij,\mathrm{loop}}
,\mathbf{v}_{ij,\mathrm{loop}})$, $(\mathbf{H}_{ji,\mathrm{loop}},\mathbf{v}_{ji,\mathrm{loop}})$
during the estimation of $\tilde{\mathbf{S}}^i_{j,\mathrm{loop}}$.

For GNSS measurements, the residual is described as follows:
\begin{equation}
\mathbf{r}_g (\mathcal{X}_i) =\mathbf{T}^n_w\circ \mathbf{T}_i\circ\mathbf{T}^b_c\circ\mathbf{t}^c_g - \tilde{\mathbf{t}}^n_{g,i}
\end{equation}
where $\tilde{\mathbf{t}}^n_{g,i}$ is the measured position of the GNSS phase center in the navigation frame, $\mathbf{t}^b_g$ is the IMU-GNSS lever-arm, $\mathbf{T}^n_w$ is a fixed world-to-navigation transformation obtained through the initial alignment.
To overcome the time asynchrony between GNSS measurements and keyframes, we introduce a temporary 
pre-integration term to bridge the GNSS measurement with the nearest keyframe timestamp. Thus, the combined cost function will be:
\begin{equation}
\left\Vert \mathbf{r}_g (\mathcal{X}_{i,\mathrm{sync}}) \right\Vert^2 + \left\Vert \mathbf{r}_b (\mathcal{X}_i,\mathcal{X}_{i,\mathrm{sync}}) \right\Vert^2 
\end{equation}
where $\mathcal{X}_{i,\mathrm{sync}}$ is a temporary node extended from $\mathcal{X}_{i}$ based on IMU data.

During global optimization, due to the inconsistency between odometry poses and loop closure relative poses, 
directly using the Hessian-form loop information for optimization is instable (since the initial estimate deviates from 
the linearization point when estimating $\tilde{\mathbf{S}}^i_{j,\mathrm{loop}}$). Therefore, in the global pose graph, we first introduce loop closure constraints in the 
form of relative poses (\ref{relative}). The resulting optimization problem is formulated as follows:
\begin{equation}
\begin{aligned}
  &\min \; \sum_{i \in \mathcal{K}}
  \left\Vert \mathbf{r}_{\mathrm{b}}(\mathcal{X}_{i}, \mathcal{X}_{i+1}) \right\Vert^2
  + \sum_{(i,j) \in \mathcal{E}}
    \mathbf{E}_{\mathrm{v}}(\mathcal{X}_{i}, \mathcal{X}_{j}) \\
  &+ \sum_{(i,j) \in \mathcal{L}}
  \rho_C \!\left(
    \left\Vert \mathbf{r}_{\mathrm{r}}(\mathcal{X}_{i}, \mathcal{X}_{j}) \right\Vert^2
  \right) \\
  &+ \sum_{i \in \mathcal{K}}{
  \left\Vert \mathbf{r}_{\mathrm{b}}(\mathcal{X}_{i}, \mathcal{X}_{i,\mathrm{sync}}) \right\Vert^2
  + 
    \rho_C \!\left(
    \left\Vert \mathbf{r}_{\mathrm{g}}(\mathcal{X}_{i,\mathrm{sync}}) \right\Vert^2
    \right)}
\end{aligned}
\end{equation}
where $\mathcal{L}$ is the loop closure set. 
In the above optimization process, a Cauchy robust kernel function $\rho_{C}({\cdot})$ is applied to the loop closure constraints to 
mitigate the influence of incorrect loop closures. After obtaining a reliable global pose estimate, we 
convert the inlier loop closure constraints into the Hessian form and optimize the following more fundamental 
optimization problem as follows:
\begin{equation}
\begin{aligned}
  &\min \; \sum_{i \in \mathcal{K}}
  \left\Vert \mathbf{r}_{\mathrm{b}}(\mathcal{X}_{i}, \mathcal{X}_{i+1}) \right\Vert^2
  + \sum_{(i,j) \in \mathcal{E}}
    \mathbf{E}_{\mathrm{v}}(\mathcal{X}_{i}, \mathcal{X}_{j}) \\
  &+ \sum_{(i,j) \in \mathcal{L'}}
    \mathbf{E}_{\mathrm{v}}(\mathcal{X}_{i}, \mathcal{X}_{j}) \\
    &+ \sum_{i \in \mathcal{K}}{
    \left\Vert \mathbf{r}_{\mathrm{b}}(\mathcal{X}_{i}, \mathcal{X}_{i,\mathrm{sync}}) \right\Vert^2
    + 
      \rho_C \!\left(
      \left\Vert \mathbf{r}_{\mathrm{g}}(\mathcal{X}_{i,\mathrm{sync}}) \right\Vert^2
      \right)}
\end{aligned}
\end{equation}
where $\mathcal{L'}$ is the inlier loop closure set,  the Hessian-form loop closure constraints are derived from $(\mathbf{H}_{ij,\mathrm{loop}}
,\mathbf{v}_{ij,\mathrm{loop}})$ and $(\mathbf{H}_{ji,\mathrm{loop}},\mathbf{v}_{ji,\mathrm{loop}})$. 

It is noted that during global optimization, the Hessian information $(\mathbf{H}_{ij},\mathbf{v}_{ij})$ 
with respect to the relative transformation $\mathbf{S}^i_j$ in Eq. (\ref{transformation}) 
doesn't need to be re-computed through 
performing the dense projection Eq. (\ref{residual}), 
as the relative transformation is stable.  Only the relative-to-absolute
linear mapping (Eq. (\ref{transformation})) needs to be iteratively performed, which makes the optimization highly efficient while
keep the accuracy of the visual information.

Through the stepwise optimization strategy and the precise utilization of measurement information, 
the global optimization problem can be solved efficiently while naturally  mitigating
 outliers in GNSS and loop closure measurements in a probabilistic manner, thereby achieving nearly optimal
 estimation results. In contrast, traditional global fusion methods 
 based solely on pose graphs discard the original IMU and visual information, causing inaccurate 
 VIO pose estimates to significantly affect the global estimation, particularly with respect to 
 the scale uncertainty.

 \section{Experiments}
	
 In the experimental section, we first evaluate the visual-inertial integration 
 performance of the system  
 on the public KITTI-360\cite{liao_kitti-360_2022} and SubT-MRS\cite{zhao_subt-mrs_2024} datasets, including both the 
 real-time SLAM and global optimization stages. Subsequently, we assess the performance of the 
 visual-inertial-GNSS fusion on our self-collected Wuhan urban dataset. Monocular camera information
 is used for all the tests.

 \subsection{KITTI-360 Dataset}
 KITTI-360\cite{liao_kitti-360_2022} is a multi-sensor dataset for autonomous-driving-related benchmarking,
 which consists of multiple relatively long (kilometer-level) data sequences. 
 The data scenarios include residential areas and highways. 
 We mainly utilize the front-facing monocular camera and the IMU data for the test.

 Firstly, we test the performance of real-time VIO. For comparison, multiple state-of-the-art monocular VIO schemes
 are considered, including feature-based methods (VINS-Fusion\cite{qin_vins-fusion_2020}, 
 ORB-SLAM3\cite{campos_orb-slam3_2021}), direct methods (DM-VIO\cite{stumberg_dm-vio_2022}) 
 and learning-based methods (DBA-Fusion\cite{zhou_dba-fusion_2024}). For extra reference, we also test the visual-only MASt3R-SLAM\cite{murai_mast3r-slam_2025}.

 \begin{figure}[t]
  \centering
  \subfloat[temporal tracking]{%
  \includegraphics[width=7.4cm]{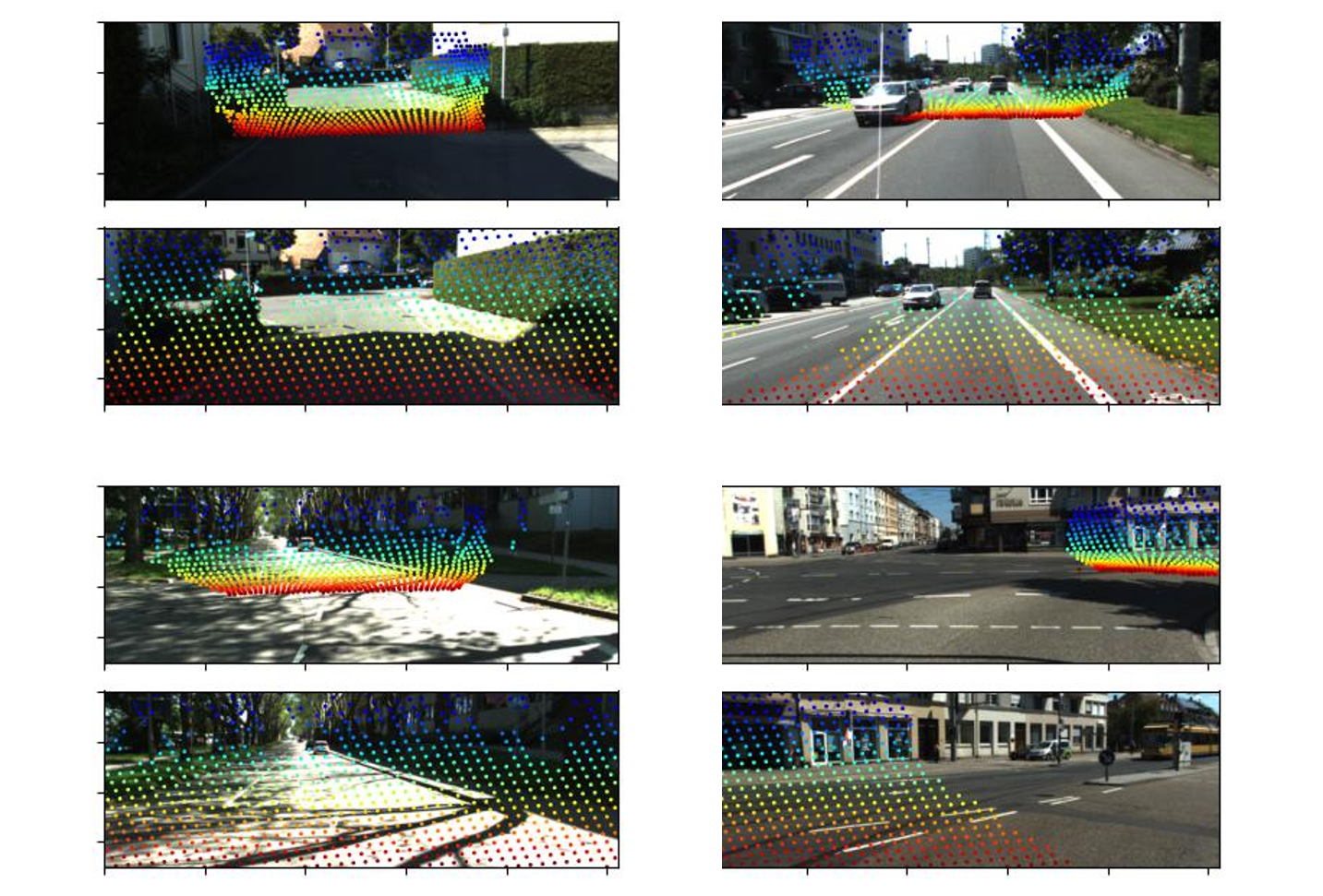}%
    \label{fig:ellipsoid_west}}\\
  \subfloat[cross-temporal matching]{%
  \includegraphics[width=6.4cm]{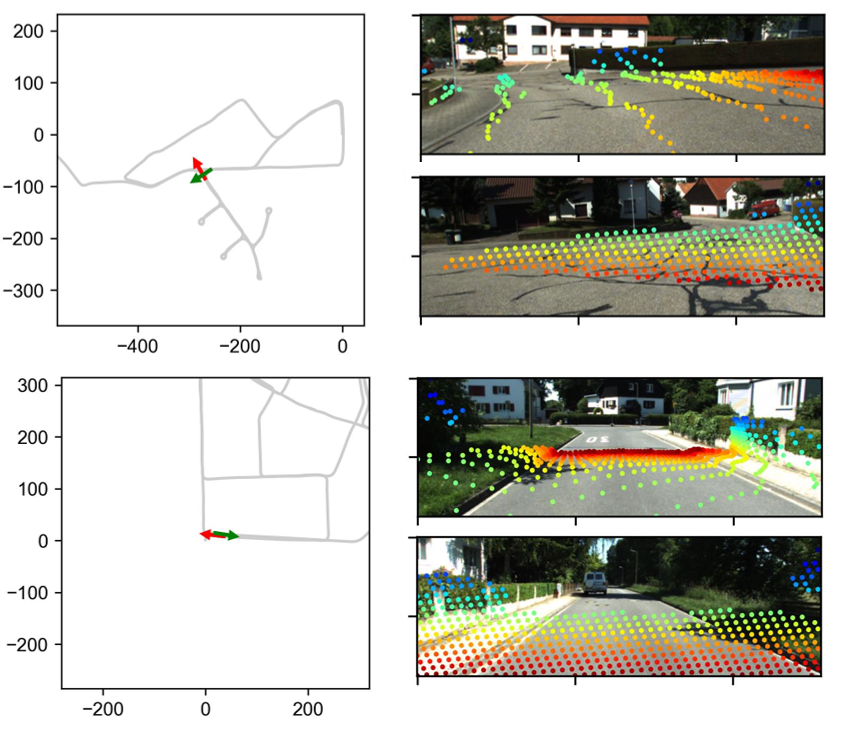}%
    \label{CDF}}
    \caption{Pixel-level association for image pairs on KITTI-360 sequences.}
    \label{fig_data_kitti}
\end{figure}

  \begin{figure}[t]
  \centering
  \subfloat[w/o loop closure]{%
  \includegraphics[width=8.3cm]{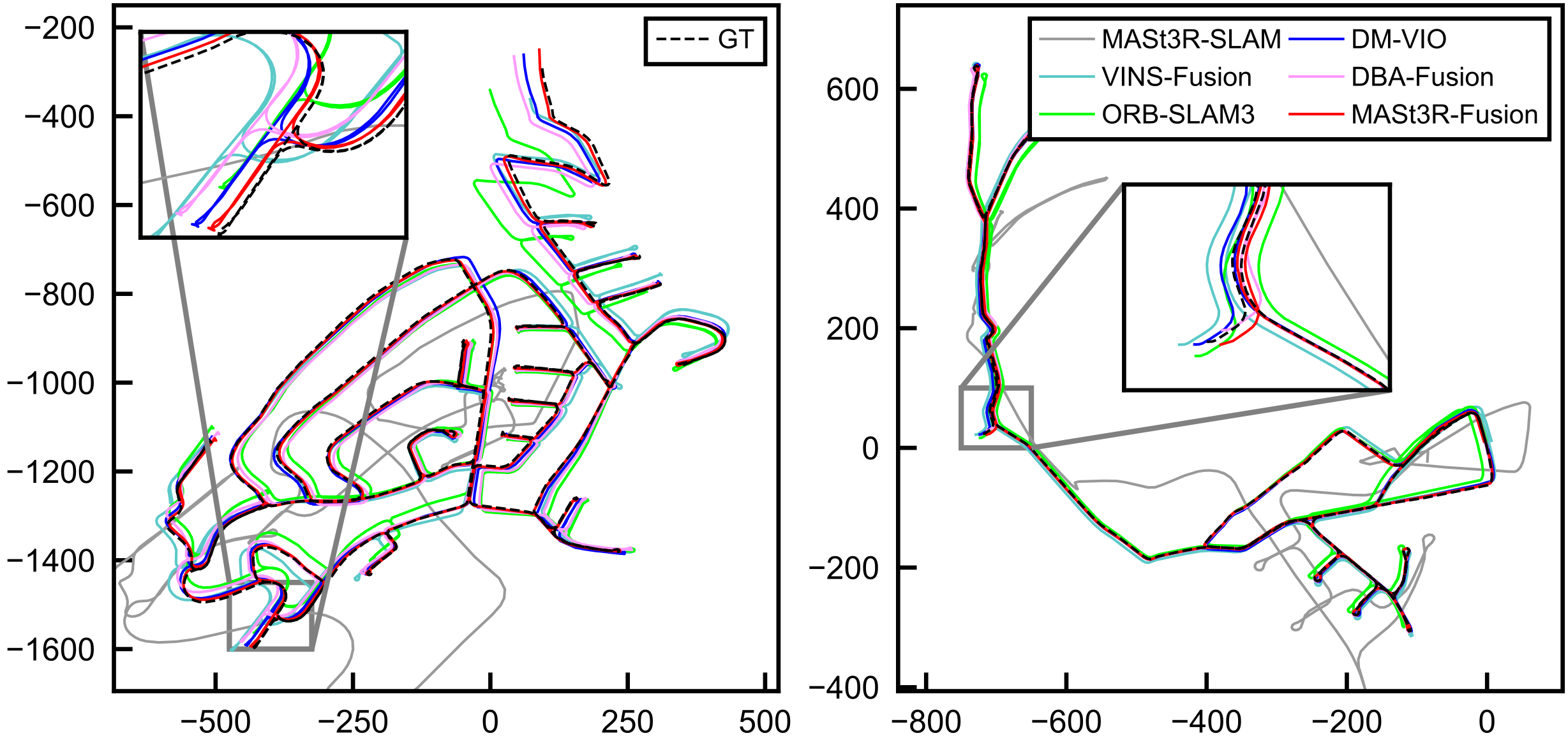}%
    \label{fig:ellipsoid_west}}\\
  \subfloat[w/ loop closure]{%
  \includegraphics[width=8.3cm]{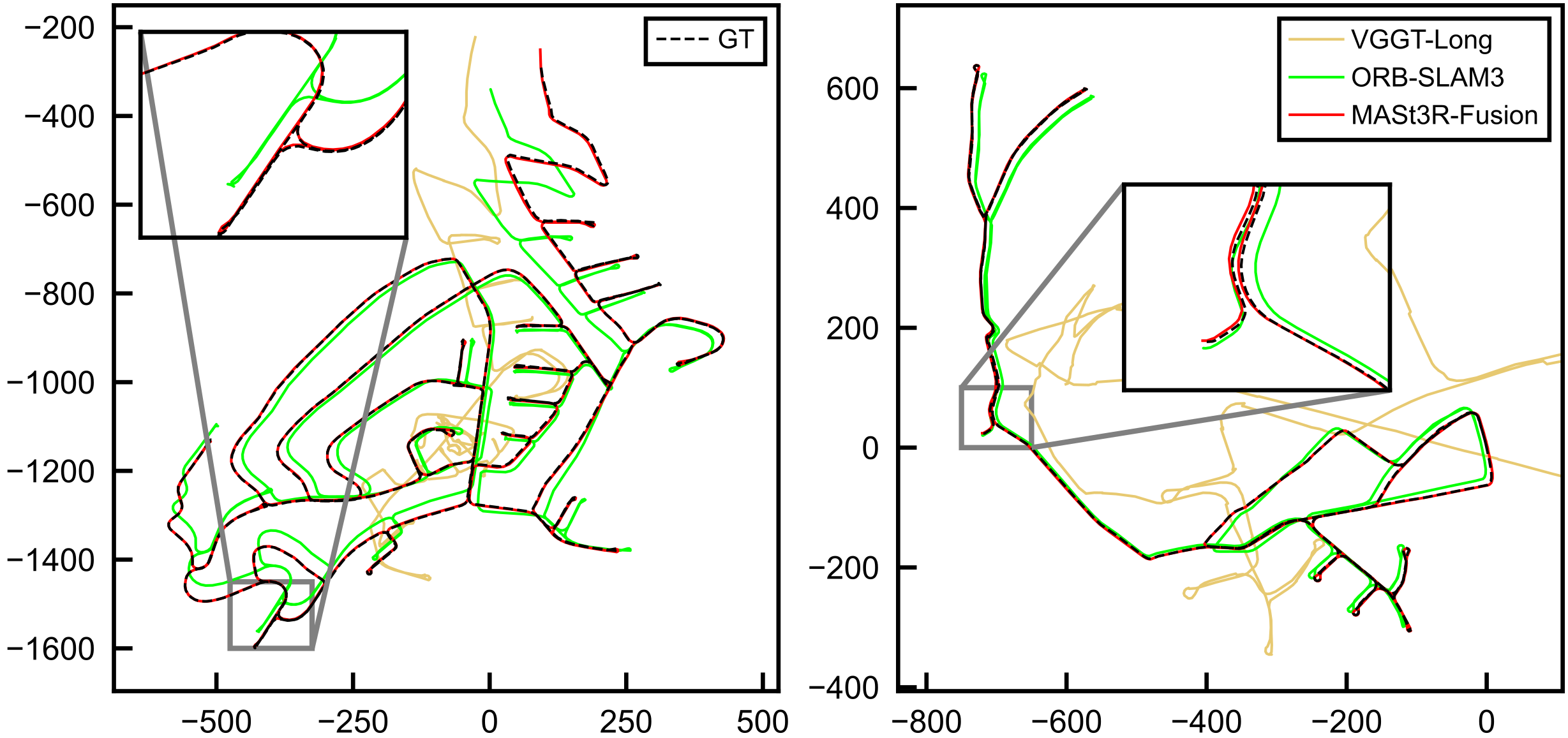}%
    \label{CDF}}
  \caption{Estimated trajectories of different V-I SLAM schemes on KITTI-360 sequences.}
  \label{fig_traj_kitti}
\end{figure}

 We first focus on the performance of data association. Fig. \ref{fig_data_kitti} shows the performance of 
 temporal tracking and cross-temporal matching based on the feed-forward model. For temporal tracking, 
 it can be observed that the feed-forward model provides dense pixel-level associations and is able 
 to directly establish reliable correspondences across frames with large intervals. Compared 
 with traditional multi-frame feature tracking methods, such long-span  associations 
 provide
  directer and stronger constraints. Besides, moving objects (e.g., cars) can be effectively excluded in the 
 association based on 3D awareness. Regarding cross-temporal matching (for loop closure), the  
 3D priors embedded in the feed-forward model makes dense matching possible even under extreme 
 viewpoint differences. To be specific, it enables imaginative data associations in cases where the viewpoint 
 difference exceeds 90°, or even when the views are completely opposite—something that is 
 difficult to achieve with conventional feature-based matching methods.

 The focus is then put on pose estimation. Considering the relatively large scale, we evaluate the odometry performance using relative pose metrics, 
 following the KITTI\cite{geiger_are_2012} odometry benchmark.
 The results are listed in Table \ref{table_kitti360}. 
 The corresponding  trajectories of representative data sequences are
  shown in Fig. \ref{fig_traj_kitti}(a). 
 It can be seen that the proposed MASt3R-Fusion achieves  accurate pose estimation, yielding lower 
 relative translation errors (RTEs) compared with both traditional indirect/direct VIO methods and 
 learning-based VIO approaches.
 The average RTE is 43.0\% lower than DM-VIO and 17.7\% lower than DBA-Fusion, 
 which shows stable performance even for weak-excitation sequences (e.g., 0003 and 0010). 
 This improvement can be attributed to the powerful data association 
 enabled by the feed-forward model, as well as the effective pointmap-alignment-based visual constraints. 
 In addition, the visual-only MASt3R-SLAM almost fails to complete SLAM at such a large scale, whose scale estimation is severely 
 affected by the imperfect information of pointmap regression. The incorporation of inertial information
  effectively mitigates these errors and fully 
 leverages the strong dense visual constraints 
 for stable metric-scale pose estimation.


 \begin{figure}[t]
   \centering 
   \includegraphics[width=8.3cm]{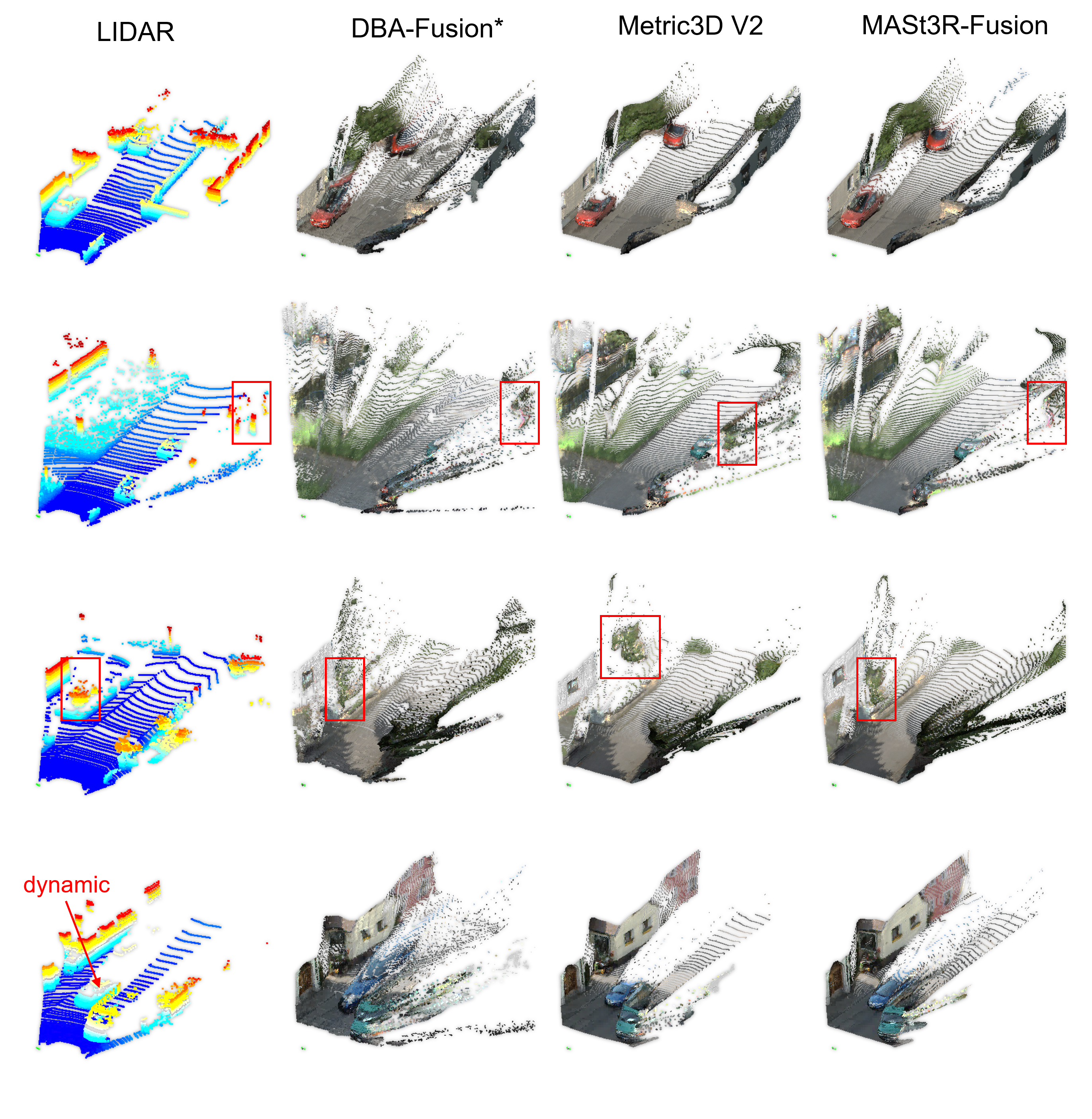}
   \caption{Real-time 3D perception performance on KITTI-360 sequences. \textsuperscript{*}For DBA-Fusion, which relies on recurrent optical flow, the perception results are obtained with a delay and are therefore not strictly real-time.} 
   \label{fig_perception} 
   \end{figure}

 \begin{table*}[h]
   \caption{Relative Pose Errors of Different VIO Schemes on KITTI-360 Dataset.}
   \label{table_kitti360}
   \setlength\tabcolsep{5pt} 
   \vspace{-10pt}
   \begin{center}
     \begin{tabular}{l|l|ll|ll|ll|ll|ll|ll}
       \hline
       \multirow{2}{*}{\makecell[c]{Seq.}} &
       \multirow{2}{*}{\makecell[c]{Desc.}} &
        \multicolumn{2}{c|}{\makecell[c]{VINS-Fusion}} & \multicolumn{2}{c|}{\makecell[c]{ORB-SLAM3}}  
        & \multicolumn{2}{c|}{\makecell[c]{DM-VIO}} & 
       \multicolumn{2}{c|}{\makecell[c]{MASt3R-SLAM\textsuperscript{*}}}  
       & \multicolumn{2}{c|}{\makecell[c]{DBA-Fusion}} &\multicolumn{2}{c}{\makecell[c]{MASt3R-Fusion}} \\ \cline{3-14} 
       &
       &\makecell[c]{ $t_{rel}$\textsuperscript{1}} & \makecell[c]{$r_{rel}$\textsuperscript{1}} 
       &\makecell[c]{ $t_{rel}$} & \makecell[c]{$r_{rel}$} 
       &\makecell[c]{ $t_{rel}$} & \makecell[c]{$r_{rel}$}
       &\makecell[c]{ $t_{rel}$} & \makecell[c]{$r_{rel}$} 
       &\makecell[c]{ $t_{rel}$} & \makecell[c]{$r_{rel}$} 
       &\makecell[c]{ $t_{rel}$} & \makecell[c]{$r_{rel}$}   \\
       \hline
       0000 & Suburb                      & \makecell[c]{1.897} & \makecell[c]{0.176} & \makecell[c]{2.386} & \makecell[c]{0.117} & \makecell[c]{1.369} & \makecell[c]{0.129}                                            & \makecell[c]{39.64} & \makecell[c]{0.524} & \makecell[c]{{\textbf{0.678}}} & \makecell[c]{{0.105}}             &           \makecell[c]{0.726} & \makecell[c]{0.151}             \\
       0002 & Suburb                      & \makecell[c]{1.006} & \makecell[c]{0.199} & \makecell[c]{1.309} & \makecell[c]{0.215} & \makecell[c]{0.724} & \makecell[c]{0.183}                                            & \makecell[c]{43.88} & \makecell[c]{0.600} & \makecell[c]{{0.577}} & \makecell[c]{{0.174}}             &           \makecell[c]{\textbf{0.504}} & \makecell[c]{0.145}             \\
       0003 & Highway                     & \makecell[c]{2.754} & \makecell[c]{{0.088}} & \makecell[c]{7.044} & \makecell[c]{0.151} & \makecell[c]{1.146} & \makecell[c]{0.111}                                          & \makecell[c]{21.59} & \makecell[c]{0.488} & \makecell[c]{{1.041}} & \makecell[c]{0.114}                      & \makecell[c]{\textbf{0.406}} & \makecell[c]{0.079}                       \\
       0004 & Suburb                      & \makecell[c]{1.710} & \makecell[c]{0.193} & \makecell[c]{1.976} & \makecell[c]{0.211} & \makecell[c]{1.063} & \makecell[c]{0.178}                                            & \makecell[c]{48.79} & \makecell[c]{0.887} & \makecell[c]{\textbf{0.556}} & \makecell[c]{0.153}   & \makecell[c]{0.770} & \makecell[c]{0.157}\\
       0005 & Suburb                      & \makecell[c]{1.187} & \makecell[c]{0.219} & \makecell[c]{1.414} & \makecell[c]{0.227} & \makecell[c]{0.729} & \makecell[c]{0.224}                                            & \makecell[c]{32.50} & \makecell[c]{0.799} & \makecell[c]{{0.619}} & \makecell[c]{{0.209}}              &   \makecell[c]{\textbf{0.544}} & \makecell[c]{0.208}                     \\
       0006 & Suburb                      & \makecell[c]{1.349} & \makecell[c]{0.176} & \makecell[c]{1.685} & \makecell[c]{0.184} & \makecell[c]{0.887} & \makecell[c]{{0.161}}                                          & \makecell[c]{54.09} & \makecell[c]{0.695} & \makecell[c]{{0.734}} & \makecell[c]{0.166}                      & \makecell[c]{\textbf{0.658}} & \makecell[c]{0.155}                       \\
       0009 & Suburb & \makecell[c]{1.596} & \makecell[c]{0.144} & \makecell[c]{2.407} & \makecell[c]{0.184} & \makecell[c]{1.379} & \makecell[c]{0.136}                                            & \makecell[c]{54.58} & \makecell[c]{0.570} & \makecell[c]{{0.846}} & \makecell[c]{{0.136}}              & \makecell[c]{\textbf{0.631}}    & \makecell[c]{0.144}                    \\
       0010 & Boulevard                   & \makecell[c]{3.610} & \makecell[c]{0.216} & \makecell[c]{5.335} & \makecell[c]{0.214} & \makecell[c]{2.130} & \makecell[c]{0.215}                                            & \makecell[c]{47.06} & \makecell[c]{0.513} & \makecell[c]{{1.486}} & \makecell[c]{{0.208}}              & \makecell[c]{\textbf{1.138}} & \makecell[c]{0.198}                       \\
       \hline
     \end{tabular}
   \end{center}
 \vspace{-5pt}
   \begin{tablenotes}
    \footnotesize
    \raggedright
   \item[*] Visual-only, scaled using Sim(3)-based global alignment\cite{grupp_evo_2017}. \item[1] $t_{rel}$ in \%, $r_{rel}$ in $^\circ$/100 m.	 
   \end{tablenotes}
 \end{table*}

 \begin{table}[h]
   \caption{Absolute Translation Errors (m) of Different 
   Global SLAM Schemes (with Loop Closure) on KITTI-360 Dataset. 
   }
   \label{table_kitti_global}
   \setlength\tabcolsep{4pt} 
   \begin{center}
   \begin{tabular}{l|lll|l}
     \hline
     \makecell[c]{Seq.} &
     \makecell[c]{VGGT\\-Long\textsuperscript{*}} &
     \makecell[c]{ORB\\-SLAM3} &
     \makecell[c]{MASt3R\\-Fusion} & 
     \makecell[c]{Leng.\\ (m)}   \\ 
     \hline
     0000 & \makecell[c]{103.64}    & \makecell[c]{26.03}       &  \makecell[c]{\textbf{2.13}} &  8361 \\
     0002 & \makecell[c]{310.76}    & \makecell[c]{32.57}       &  \makecell[c]{\textbf{2.82}} & 11195 \\
     0003 & \makecell[c]{ 26.46}   & \makecell[c]{28.63}       &  \makecell[c]{\textbf{0.70}}  &  1368 \\
     0004 & \makecell[c]{165.67}    & \makecell[c]{42.82}       &  \makecell[c]{\textbf{4.56}} &  8614 \\
     0005 & \makecell[c]{234.48}    & \makecell[c]{10.37}       &  \makecell[c]{\textbf{1.28}} &  4561 \\
     0006 & \makecell[c]{179.95}    & \makecell[c]{ 9.51}       &  \makecell[c]{\textbf{2.52}} &  7699 \\
     0009 & \makecell[c]{135.23}    & \makecell[c]{ 6.95}       &  \makecell[c]{\textbf{1.90}} &  8677 \\
     0010 & \makecell[c]{211.54}    & \makecell[c]{45.17}       &  \makecell[c]{\textbf{4.38}} &  3340 \\
     \hline
     ave(\%)& \makecell[c]{2.91}& \makecell[c]{0.63} & \makecell[c]{\textbf{0.05}} & \makecell[c]{norm.} \\
     \hline
   \end{tabular}
 \end{center}
 \vspace{-5pt}
   \begin{tablenotes}
    \footnotesize
    \raggedright
     \item[*] Visual-only, scaled using Sim(3)-based global alignment\cite{grupp_evo_2017}.
   \end{tablenotes}
 \end{table}

 Afterwards, we evaluate the global SLAM performance by 
 activating loop closure and global optimization.
 For comparison, we test ORB-SLAM3 which supports loop closure and BA-based global optimization, 
 and VGGT-Long\cite{deng_vggt-long_2025} which utilizes state-of-the-art feed forward model\cite{wang_vggt_2025} 
 and chunked processing strategy
 for long-sequence  reconstruction. We simply
 account the absolute translation errors (ATEs) for evaluation\cite{grupp_evo_2017}, as listed in Table \ref{table_kitti_global}.
The corresponding trajectory is shown in Fig. \ref{fig_traj_kitti}(b). It can be observed that the proposed 
MASt3R-Fusion achieves 
a significantly lower ATE than ORB-SLAM3.
The average ATE/length achieves 0.05\%, which is quite impressive considering that only monocular V-I data is used.
This is partly due to the higher odometry accuracy, 
and partly because the proposed method incorporates stronger and richer loop closure information. 
As illustrated in Fig. \ref{fig_data_kitti}, loop closures under large viewpoint differences 
allow the system to maintain pose consistency even when driving in opposite directions, 
which provides favorable conditions for globally consistent mapping. As for the visual-only VGGT-Long, although
the multi-view feed-forward  model is employed,
its effectiveness is somewhat limited. The significant scale drift of the visual system in large-scale scenes
makes the global  optimization challenging, and makes it sensitive to false loop 
closures, yielding to conservative strategies. In contrast, the 
incorporation of IMU inforamtion in MASt3R-Fusion  allows the potential of feed-forward visual models to be fully realized.

 One of the functionalities of the proposed system is real-time 3D spatial percpetion with metric scale awareness.
 In Fig. \ref{fig_perception}, we qualitatively evaluate this 
 by comparing the proposed system with learning-based dense VIO\cite{zhou_dba-fusion_2024} system 
 and metric depth inference model\cite{hu_metric3d_2024}.
 From the results, it can be observed that all the three methods are capable of achieving dense 
 perception of the environment with mertic scale. Among them, DBA-Fusion primarily 
 relies on BA for reconstructing the environment structure but lacks 3D priors, 
 which leads to scattered points and poor handling of dynamic objects in the scene. 
 The monocular depth model Metric3D v2 (ViT-Large), through specialized training, demonstrates strong depth estimation performance, particularly in terms of surface normals and edges.
  However, since it fully depends on inference, objects with ambiguous depth cues can
 exhibit significant depth/scale errors. In contrast, MASt3R-Fusion preserves the overall 3D structure 
 while leveraging binocular geometry, thereby attaining relatively 
 stable and complete real-time 3D  perception performance.

 \begin{figure}[t]
   \centering
   \subfloat[temporal tracking]{%
   \includegraphics[width=8.3cm]{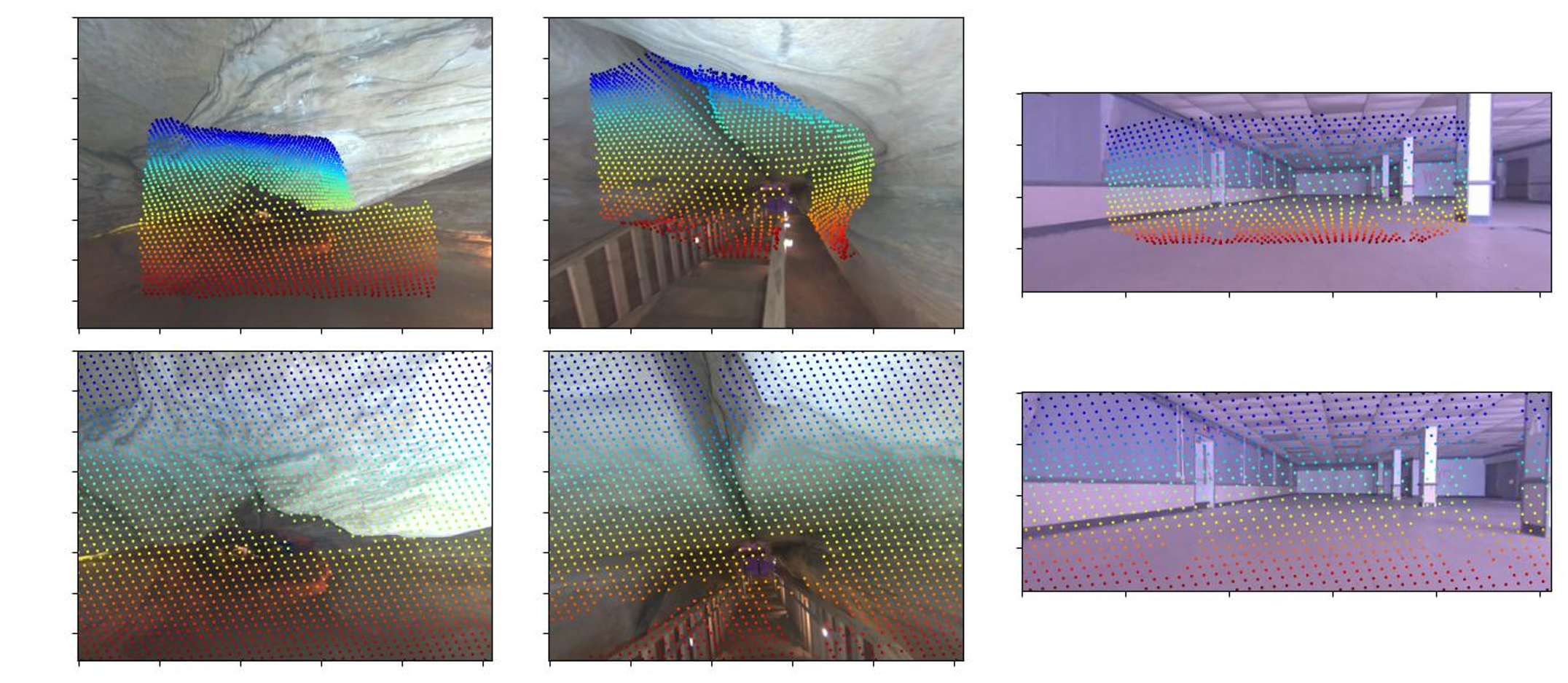}%
     \label{fig:ellipsoid_west}}\\
   \subfloat[cross-temporal matching]{%
   \includegraphics[width=6.4cm]{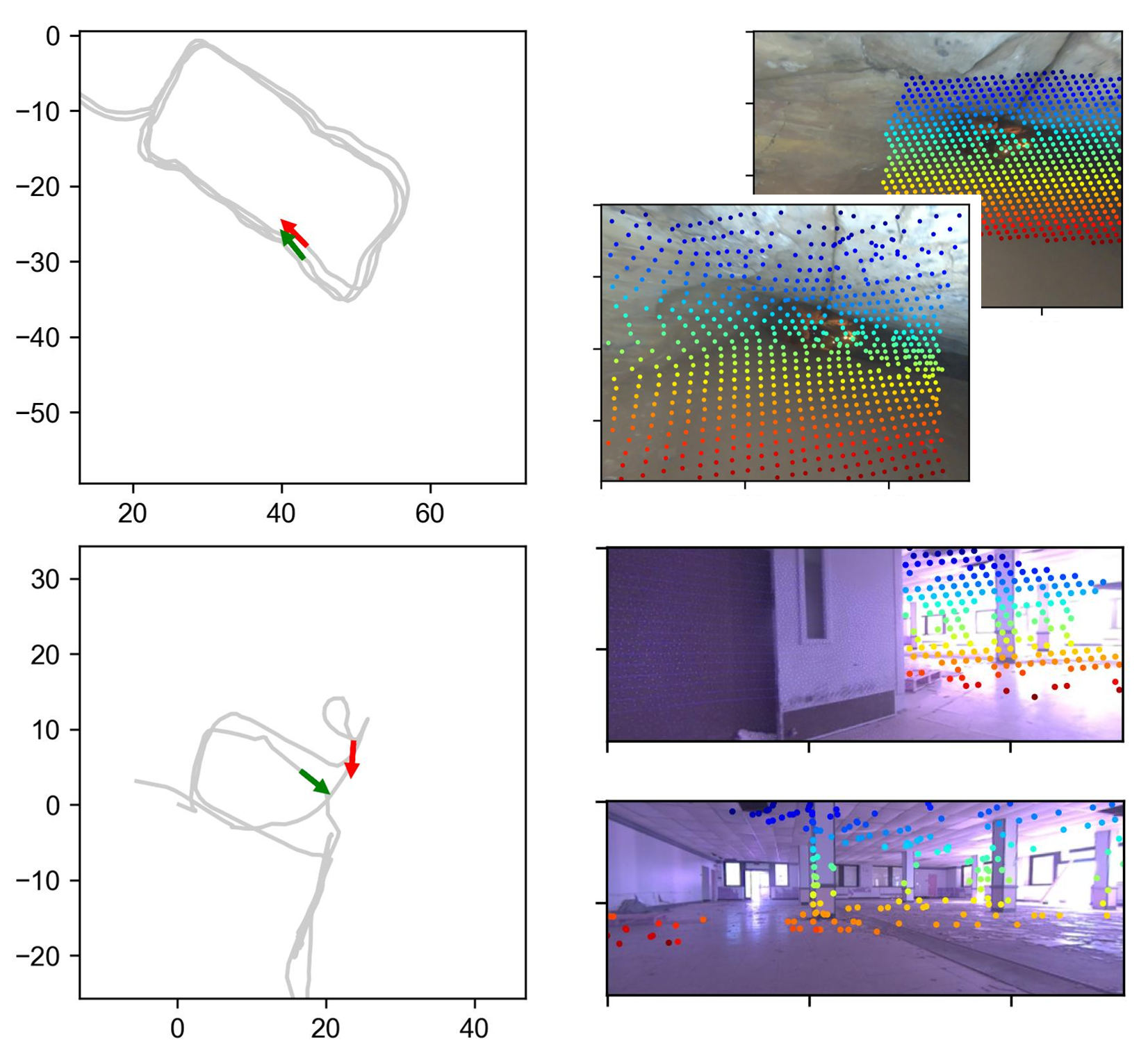}%
     \label{fig:ellipsoid_east}}
   \caption{Pixel-level association for image pairs on SuBT-MRS sequences.}
   \label{subt}
 \end{figure}

 \begin{figure}[t]
   \centering
   \subfloat[w/o loop closure]{%
   \includegraphics[width=8.3cm]{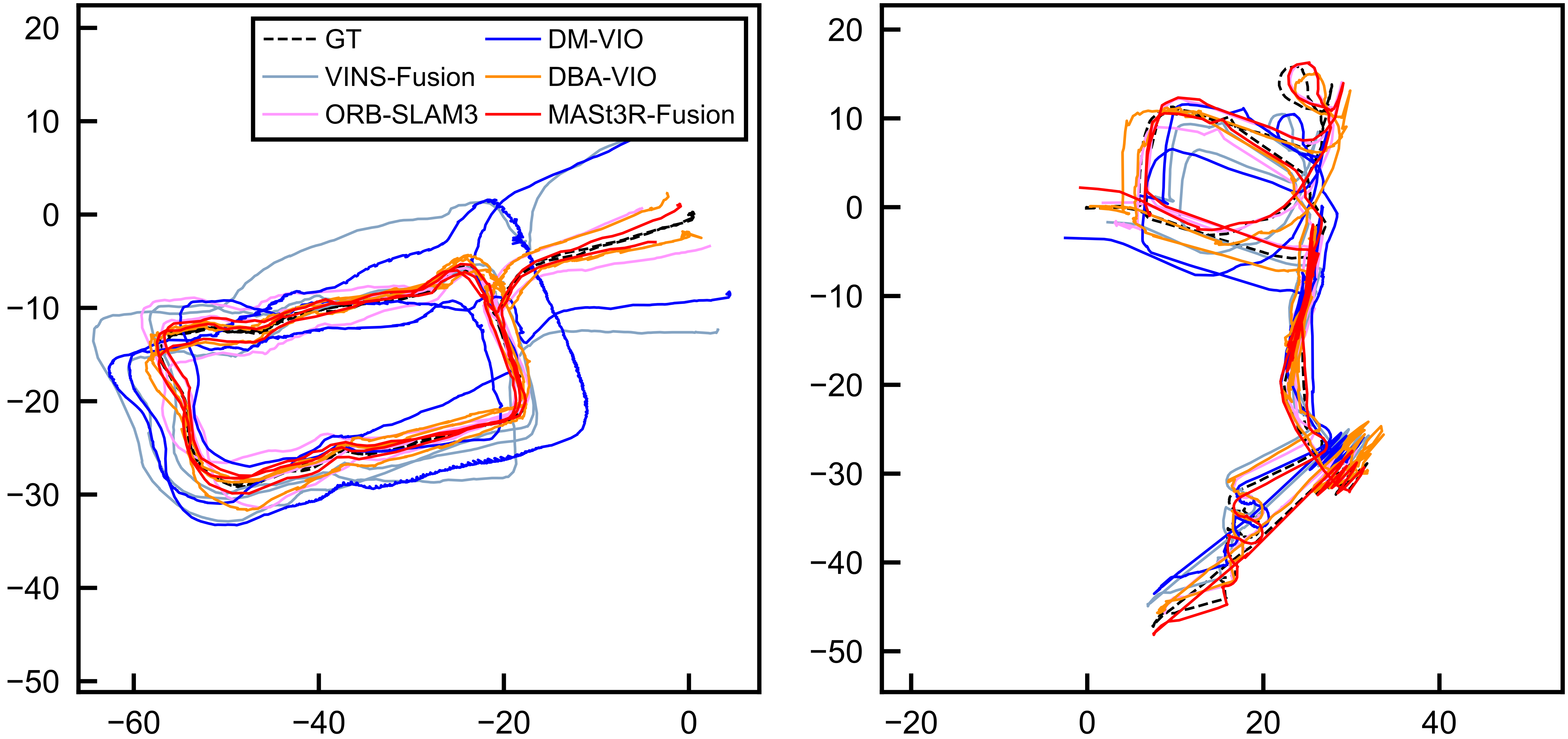}%
     \label{fig:ellipsoid_west}}\\
   \subfloat[w/ loop closure]{%
   \includegraphics[width=8.3cm]{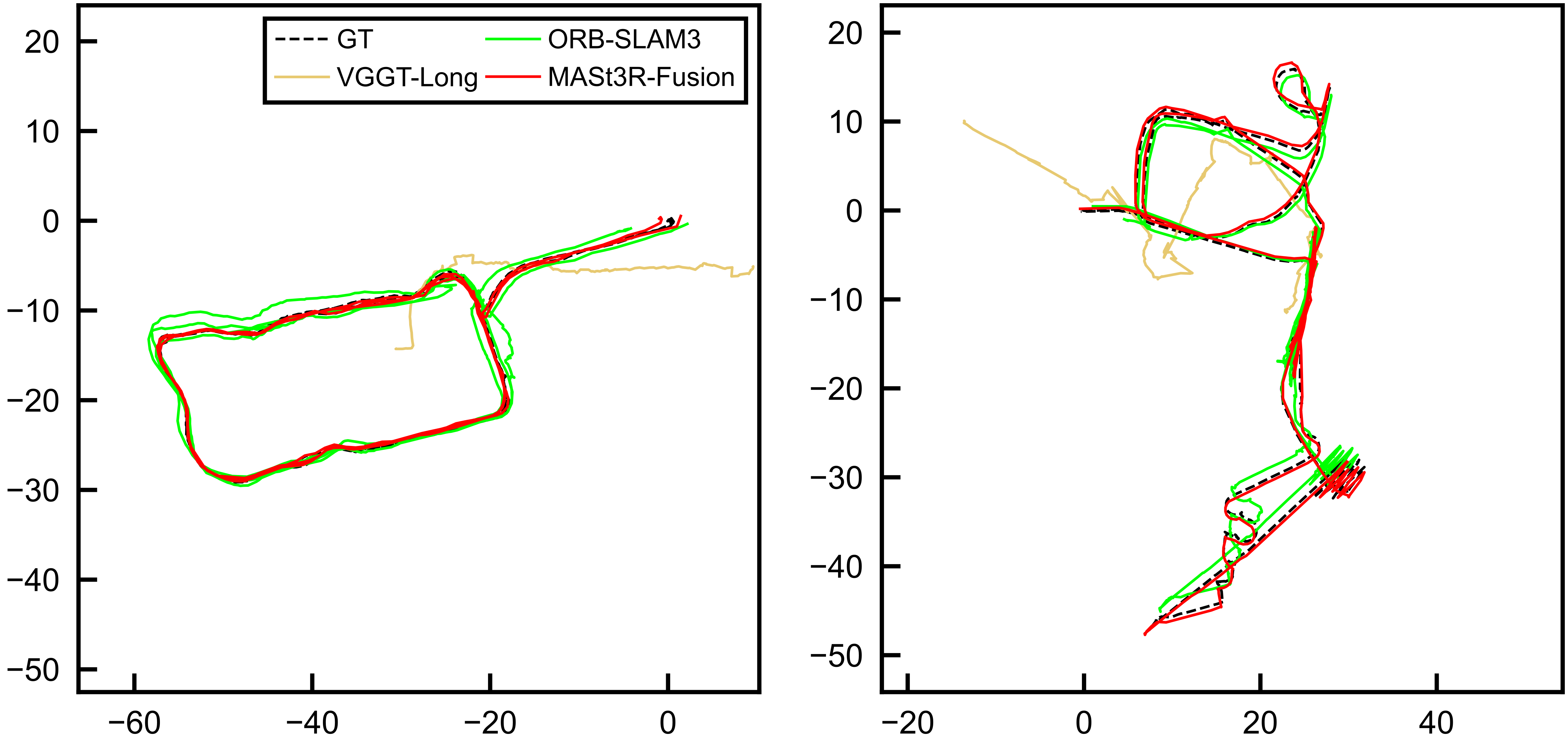}%
     \label{fig:ellipsoid_east}}
   \caption{Estimated trajectories of different V-I SLAM schemes on SuBT-MRS sequences.}
   \label{fig_subt_traj}
 \end{figure}

 \subsection{SubT-MRS Dataset}

 An important issue for general-purpose visual models is their 
 generalization ability. For example, existing monocular depth 
 estimation algorithms can perform well in rooms 
 or driving scenarios, but  may face challenges in unseen, 
 unusual scenes. To evaluate the effectiveness of the algorithm 
 in unconventional scenarios, the V-I dataset from Subt-MRS\cite{zhao_subt-mrs_2024} is employed. 
 This dataset contains three real-world sequences, two collected in 
 karst caves, and one recorded by a quadruped robot 
 transitioning between indoor and outdoor environments.

 Similarly, we first focus on the performance of data association. 
 It can be seen from Fig. \ref{subt}  that in unconventional scenarios (such as caves), the feed-forward 
 model still provides reliable dense pixel associations, even when the scene 
 texture is only weakly distinguishable. The matching performance under large 
 viewpoint differences remains excellent, even surpassing the range discernible 
 by the human eye, thereby offering strong potential constraints for global SLAM.

 Afterwards, we conducted tests on these data sequences for both real-time 
  and global V-I SLAM. The results of the pose estimation are presented in Table \ref{table_subt} and Table \ref{table_subt_global},
  and corresponding trajectories are shown in Fig. \ref{fig_subt_traj}.

 \begin{table}[h]
  \caption{Absolute Translation Errors (m) of Different VIO Schemes on SubT-MRS Dataset.
  }
  \label{table_subt}
  \setlength\tabcolsep{3pt} 
  \begin{center}
  \begin{tabular}{l|lllll|l}
    \hline
    \makecell[c]{Seq.} & 
    \makecell[c]{VINS} & 
    \makecell[c]{ORB} & 
    \makecell[c]{DM-VIO} & 
    \makecell[c]{DBA} & 
    \makecell[c]{M-Fus.} &
    \makecell[c]{Leng.\\ (m)}   \\ 
    \hline
    handheld1 & \makecell[c]{5.64} & \makecell[c]{2.16} & \makecell[c]{14.54} & \makecell[c]{1.78}    & \makecell[c]{\textbf{1.07}}  & \makecell[c]{394}\\
    handheld2 & \makecell[c]{2.32} & \makecell[c]{2.42} & \makecell[c]{4.94} & \makecell[c]{2.09}     & \makecell[c]{\textbf{1.13}}   & \makecell[c]{509}\\
    overexposure & \makecell[c]{2.61} & \makecell[c]{1.27} & \makecell[c]{2.97} & \makecell[c]{1.85}   & \makecell[c]{\textbf{0.99}} & \makecell[c]{509}\\
    \hline
    ave(\%) & \makecell[c]{0.80} & \makecell[c]{0.42} & \makecell[c]{1.74} & \makecell[c]{0.41}  & \makecell[c]{\textbf{0.23}}  & \makecell[c]{norm.}\\
    \hline
  \end{tabular}
\end{center}
  \begin{tablenotes}
  \end{tablenotes}
\end{table}

\begin{table}[h]
  \caption{Absolute Translation Errors (m) of Different Global SLAM Schemes (with Loop Closure) on SubT-MRS Dataset.
  }
  \label{table_subt_global}
  \setlength\tabcolsep{4pt} 
  \begin{center}
  \begin{tabular}{l|lll|l}
    \hline
    \makecell[c]{Seq.} &
    \makecell[c]{VGGT\\-Long\textsuperscript{*}} &
    \makecell[c]{ORB\\-SLAM3} &
    \makecell[c]{MASt3R\\-Fusion} & 
    \makecell[c]{Leng.\\ (m)}   \\ 
    \hline
    handheld1 &     \makecell[c]{fail}     &  \makecell[c]{1.48}     &  \makecell[c]{\textbf{0.26}} &  \makecell[c]{394} \\
    handheld2 &     \makecell[c]{fail}      & \makecell[c]{2.14}      &  \makecell[c]{\textbf{1.04}} &  \makecell[c]{395} \\
    overexporsure & \makecell[c]{fail}    &  \makecell[c]{1.07 }    &  \makecell[c]{\textbf{0.43}} &  \makecell[c]{509} \\
    \hline
    ave(\%) & \makecell[c]{-} & \makecell[c]{0.37} & \makecell[c]{\textbf{0.13}} & \makecell[c]{norm.} \\
    \hline
  \end{tabular}
\end{center}
\vspace{-5pt}
  \begin{tablenotes}
   \footnotesize
   \raggedright
    \item[*] Visual-only.
  \end{tablenotes}
\end{table}

 From the real-time SLAM results, it can be observed that MASt3R-Fusion achieves significantly 
 higher pose estimation accuracy than existing methods, providing stable metric-scale odometry performance. In terms of
  global SLAM performance, the aggressive loop closure detection  delivers impressive results, 
  leading to a substantial reduction in ATEs.
 This part of the experiments largely validates the generalizability of the feed-forward model 
 approach on open scenarios when incorporated with IMU information.

 \subsection{Wuhan Urban Dataset}
\begin{figure}[t]
   \centering
   \includegraphics[width=8.3cm]{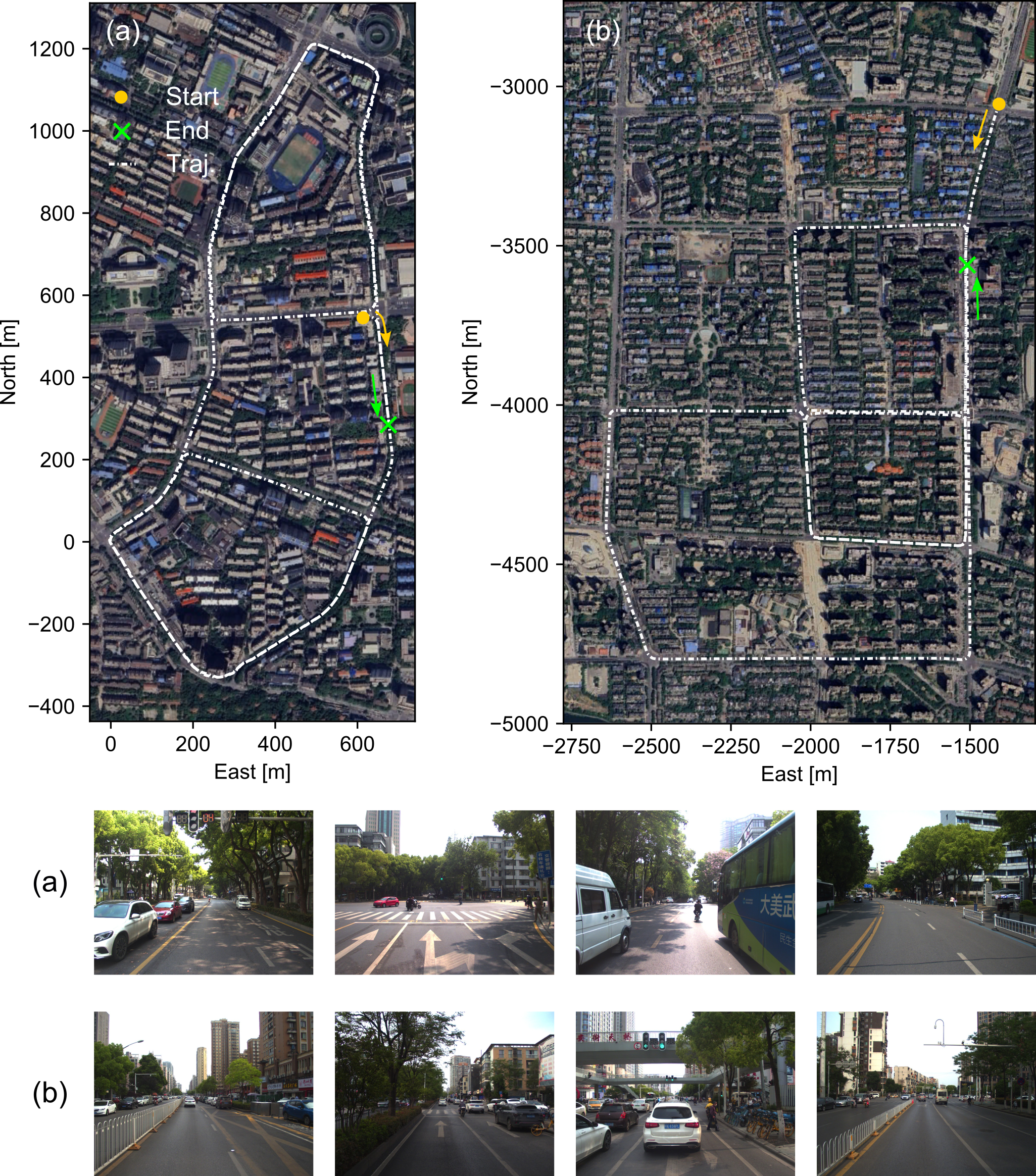}
   \caption{Trajectories and environmental images of  sequences in the self-collected Wuhan urban dataset.}
   \label{fig_wuhan_bev}
 \end{figure}

 In addition to the evaluation of the V-I system, to further evaluate the algorithm's 
 performance in real-world road scenarios (with more dynamic objects) and to consider its integration 
 with global GNSS measurements, we turn to the  self-collected multi-sensor dataset acquired in Wuhan City. The dataset 
 contains two sequences, with their trajectories shown in Fig. \ref{fig_wuhan_bev}. 
 The experimental vehicle is equipped with an RGB camera, an ADIS16470 IMU, 
 a Septentrio AsterRx4 GNSS receiver and a tactical-grade IMU for ground-truth trajectory generation.
 A nearby GNSS base station is available for differential GNSS (DGNSS) processing. We take the classical VINS-Fusion
 as the baseline, whichs supports both real-time pose estimation and global fusion 
 with GNSS. We also test DBA-Fusion which employs a learning-based visual frontend and supports real-time V-I(/GNSS) integrated navigation.

 \begin{figure}[t]
   \centering
   \includegraphics[width=8.3cm]{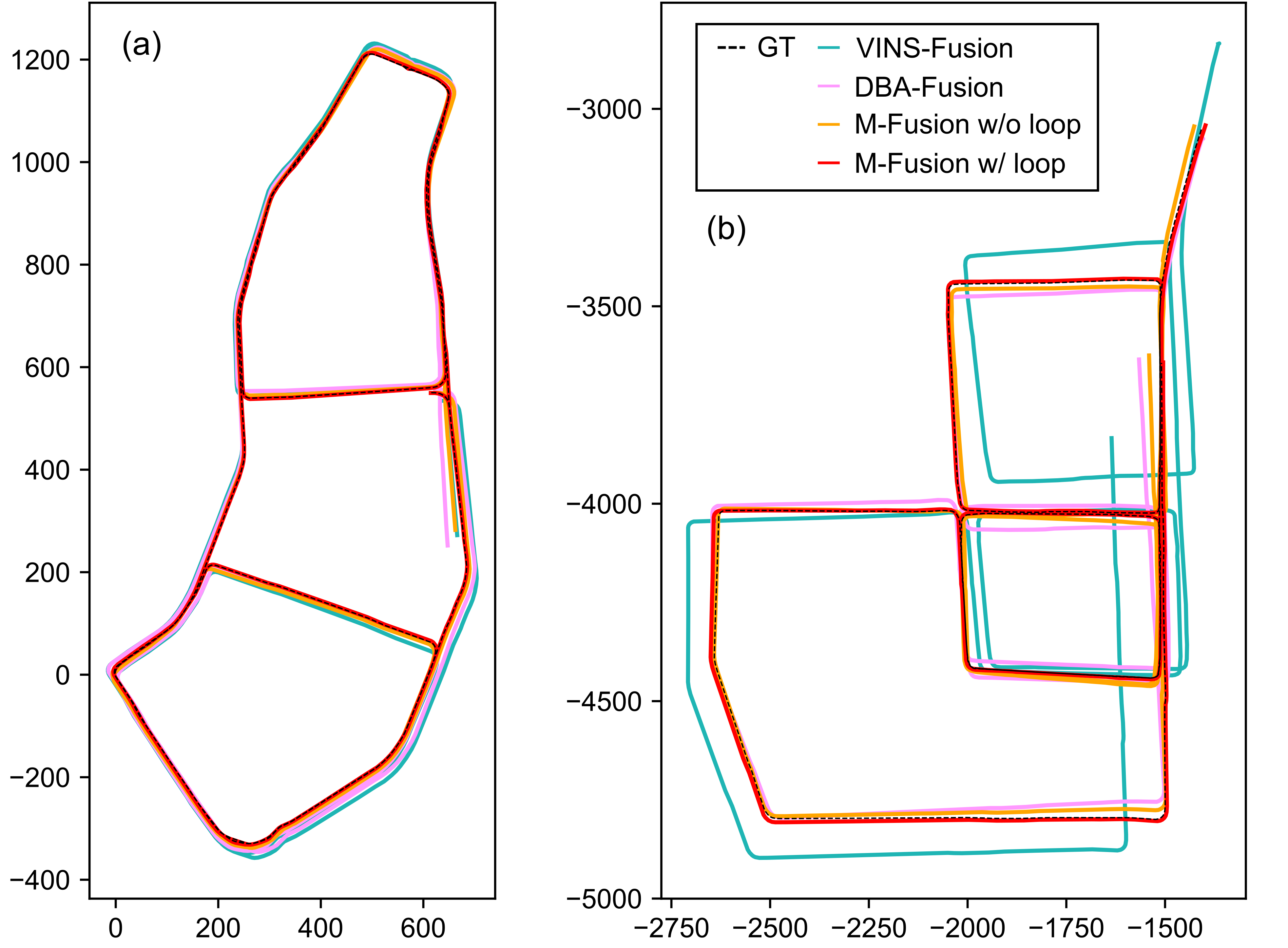}
   \caption{Estimated trajectories of different V-I SLAM schemes on Wuhan urban sequences.}
   \label{fig_wuhan_odo}
 \end{figure}

First, we tested several different pose estimation schemes based on V-I data, including real-time VIO and global SLAM. 
The corresponding trajectories are shown in Fig. \ref{fig_wuhan_odo}, with the relative pose estimation errors listed in Table \ref{table_whu_vio}. It can be observed that in large-scale road 
environments (with  long straight segments), the traditional VIO algorithm suffers from significant 
scale estimation errors due to weak observability\cite{martinelli_visual-inertial_2013}. 
This problem remains for DBA-Fusion, although the dense visual information does improve the 
VIO performance to some extent.
In contrast, the proposed MASt3R-Fusion method, by combining probabilistic 
marginalization with Sim(3)-based visual alignment constraints, can effectively 
control and 
correct the scale drift. Besides, after introducing loop closure constraints, the position error caused 
by attitude drift is greatly reduced, and the trajectory accuracy is further improved.

\begin{table}[h]
 \caption{Relative Pose Errors of Different VIO Schemes on KITTI-360 Dataset.}
 \label{table_whu_vio}
 \setlength\tabcolsep{4pt} 
 \begin{center}
   \begin{tabular}{l|ll|ll|ll|ll}
     \hline
     \multirow{2}{*}{\makecell[c]{Seq.}} &
      \multicolumn{2}{c|}{\makecell[c]{VINS\\-Fusion}} & 
     \multicolumn{2}{c|}{\makecell[c]{DBA\\-Fusion}}  
     & \multicolumn{2}{c|}{\makecell[c]{M-Fus. \\w/o loop}} &
     \multicolumn{2}{c}{\makecell[c]{M-Fus. \\w/ loop}} \\ \cline{2-9} 
     &\makecell[c]{ $t_{rel}$\textsuperscript{1}} & \makecell[c]{$r_{rel}$\textsuperscript{1}} 
     &\makecell[c]{ $t_{rel}$} & \makecell[c]{$r_{rel}$} 
     &\makecell[c]{ $t_{rel}$} & \makecell[c]{$r_{rel}$} 
     &\makecell[c]{ $t_{rel}$} & \makecell[c]{$r_{rel}$}   \\
     \hline
     (a) & \makecell[c]{2.07} & \makecell[c]{0.11} & \makecell[c]{1.45} & \makecell[c]{0.11} & \makecell[c]{1.20} & \makecell[c]{0.11}& \makecell[c]{\textbf{0.53}} & \makecell[c]{0.07}           \\
     (b) & \makecell[c]{5.85} & \makecell[c]{0.13} & \makecell[c]{3.01} & \makecell[c]{0.13} & \makecell[c]{1.68} & \makecell[c]{0.13}& \makecell[c]{\textbf{0.92}} & \makecell[c]{0.06}           \\
     \hline
   \end{tabular}
 \end{center}
 \vspace{-5pt}
 \begin{tablenotes}
  \footnotesize
  \raggedright
 \item[1] $t_{rel}$ in \%, $r_{rel}$ in $^\circ$/100 m.	 
 \end{tablenotes}
\end{table}

\begin{figure}[!t]
 \centering 
 \includegraphics[width=8.3cm]{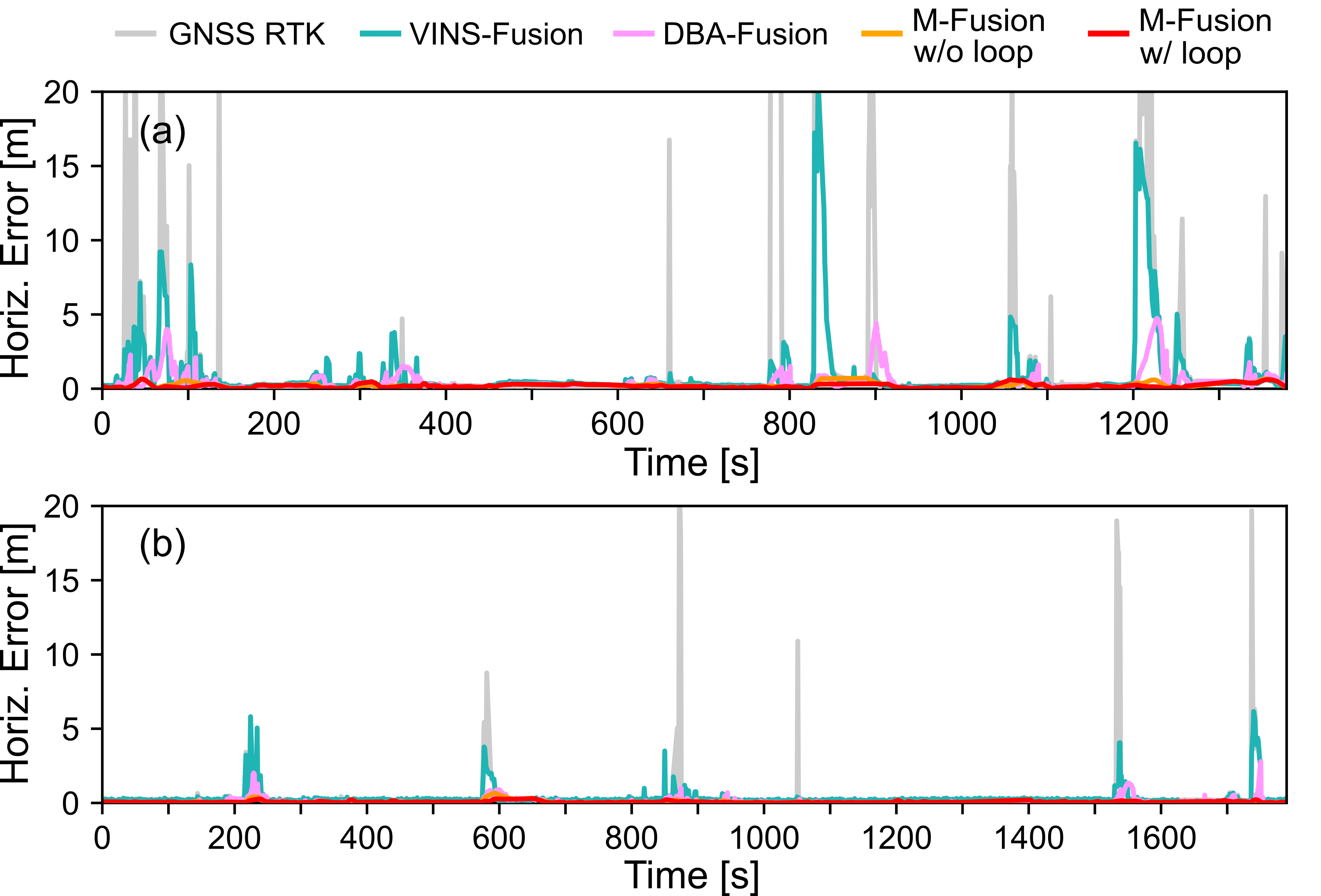}
 \caption{Horizontal errors of different absolute positioning schemes with GNSS RTK integration.
  } 
 \label{fig_gnss0} 
 \end{figure}

Second, the global positioning performance with GNSS integration is focused on. The positioning
errors are depicted in Fig. \ref{fig_gnss0} and listed in Table \ref{table_whu_global}.
From the results, it can be observed that GNSS RTK exhibits many  gross errors and periods of unavailability,
especially in sequence (a).
For VINS-Fusion, in which GNSS positions are fused with VIO-provided relative poses loosely through 
a global pose graph, the position estimation errors during GNSS-degraded periods 
cannot be effectively suppressed, since the position residuals cannot be 
probabilistically distributed across all poses. 
This further undermines the 
detection and mitigation of GNSS gross errors. For DBA-Fusion, although the GNSS inforamtion
can assist in the V-I subsystem, it only supports real-time fusion and cannot provide global optimal solutions.
 In contrast, in the proposed 
MASt3R-Fusion, the complete and precise V-I information is fully 
preserved into the factor graph, effectively resisting GNSS gross errors through iterative optimization
and maintaining decimeter-level accuracy through the period.

\begin{figure}[!t]
 \centering 
 \includegraphics[width=8.3cm]{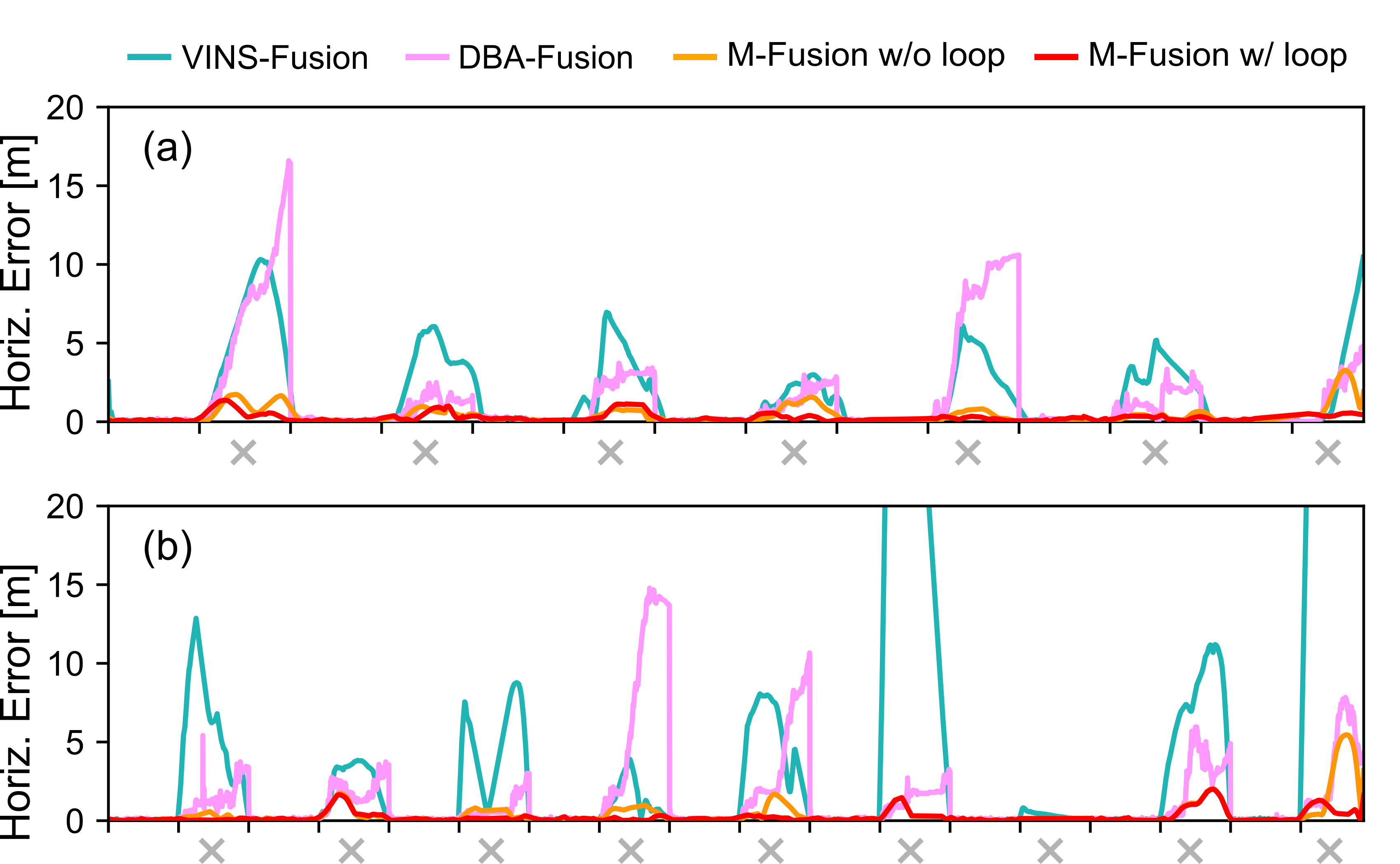}
 \caption{Horizontal errors of different absolute positioning schemes with simulated GNSS integration.
 The gray crosses indicate simulated GNSS outages (100-second). } 
 \label{fig_gnss10} 
 \end{figure}

Moreover, to enrich the variability of GNSS observation conditions in the tests, we use simulated GNSS data
based on ground truth with intermittent 
outages  to assess the robustness of the algorithm. The results are shown in 
Fig. \ref{fig_gnss10} and listed in Table \ref{table_whu_global}, which further demonstrate the effectiveness of the proposed method in
recovering globally optimal trajectories. Impressively, under  100-second GNSS outages, 
the global factor graph can still achieve mostly sub-meter level trajectory smoothing, and 
the introduction of cross-temporal associations further reduces the overall error.

 \begin{table}[h]
 \caption{Horizontal Position RMSEs (m) of Different Positioning 
 Schemes with GNSS RTK integration or with simulated GNSS integration.
 }
 \label{table_whu_global}
 \setlength\tabcolsep{4pt} 
 \begin{center}
 \begin{tabular}{l|l|lllll}
   \hline
   \makecell[c]{Mode} &
   \makecell[c]{Seq.} &
   \makecell[c]{GNSS \\ RTK} &
   \makecell[c]{VINS \\ Fusion} &
   \makecell[c]{DBA \\ Fusion} & 
   \makecell[c]{M-Fus.\\ w/o loop} & 
   \makecell[c]{M-Fus.\\ w/ loop}\\ 
   \hline
   \multirow{2}{*}{Real}  & (a) &     \makecell[c]{4.36}     &  \makecell[c]{2.54}     &  \makecell[c]{0.78} &  \makecell[c]{0.24}&  \makecell[c]{\textbf{0.21}} \\
    & (b) &     \makecell[c]{1.46}      & \makecell[c]{0.62}      &  \makecell[c]{0.24} &  \makecell[c]{0.13}&  \makecell[c]{\textbf{0.09}} \\
   \hline
   \multirow{2}{*}{Simu.}  &(a) &     \makecell[c]{-}     &  \makecell[c]{2.84}      &  \makecell[c]{3.20} &  \makecell[c]{0.69} &  \makecell[c]{\textbf{0.37}} \\
                           &(b) &     \makecell[c]{-}     &  \makecell[c]{9.66}      &  \makecell[c]{2.94} &  \makecell[c]{1.02} &  \makecell[c]{\textbf{0.46}} \\
   \hline
 \end{tabular}
\end{center}
 \begin{tablenotes}
 \end{tablenotes}
\end{table}

\begin{figure*}[!t]
 \centering 
 \includegraphics[width=17.4cm]{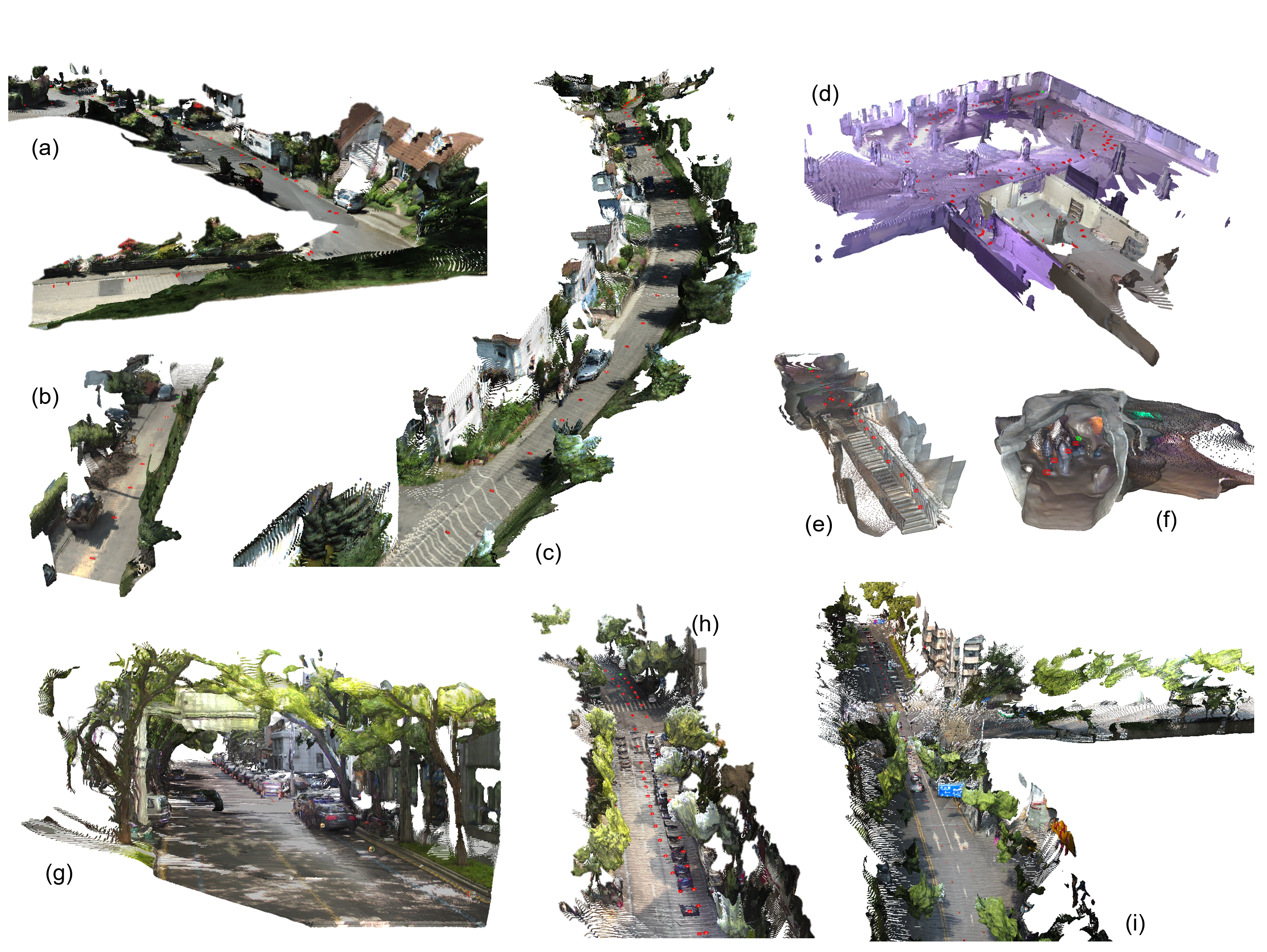}
 \caption{Reconstruction results. (a-c) KITTI-360. (d-f) SubT-MRS. (g-i) Self-collected Wuhan urban dataset.} 
 \label{fig_zuizhong} 
 \end{figure*}

 \subsection{Reconstruction Results}
 
 Finally, we qualitatively evaluate the dense 3D perception 
 and scene reconstruction performance of the proposed method under different scenarios. 
 As shown in Fig. \ref{fig_zuizhong}, the proposed method can effectively recover 
 the dense structure of the scene with metric scale. 
 For revisited trajectories, the consistency of scene reconstruction can
 be achieved, which based on the high-accuracy pose estimation and effective cross-temporal data association.

	\section{Conclusion}
	In this work, we presented \textit{MASt3R-Fusion}, a novel multi-sensor-assisted
	 visual SLAM framework that tightly integrates feed-forward pointmap regression with 
	 complementary sensor information, including inertial and GNSS measurements. 
	 By introducing Sim(3)-based visual alignment constraints into a 
	 universal metric-scale SE(3) factor graph, our system achieves accurate pose tracking, metric-scale 
	 3D structure perception and globally consistent SLAM.
 Experimental results on both public benchmarks and self-collected datasets indicate the  
 substantial improvements over existing visual and multi-sensor SLAM systems, showing  high
 functionality in terms of localization and mapping.
	
 This work facilitates robust and accurate pose estimation and mapping in 
 ubiquitous scenarios. In the future, our research 
 will focus on semantic fusion and  advanced scene representations to support embodied navigation tasks.

	\section*{Acknowledgments}
	This work was supported by the National Science Fund for
			Distinguished Young Scholars of China (42425401), the National
			Natural Science Foundation of China (423B240), the
			National Key Research and Development Program of China 
      (2023YFB3907100), and the China Postdoctoral Science Foundation-Hubei Joint 
      Support Program (2025T041HB).
		The implemented MASt3R-Fusion is developed by the GREAT Group, School of Geodesy and Geomatics, Wuhan University. The numerical calculations
	in this paper have been done on the supercomputing system
	at the Supercomputing Center of Wuhan University.
\bibliographystyle{IEEEtran}
	\bibliography{exported}
	
\end{document}